\documentclass[journal]{IEEEtran}
\usepackage{amsmath,amssymb,bm}
\usepackage{array,threeparttable,booktabs,multirow}
\usepackage{graphicx}
\usepackage[noend]{algpseudocode}
\usepackage{cite}
\usepackage[dvipsnames]{xcolor}
\usepackage{setspace}
\usepackage[colorlinks=true, allcolors=blue]{hyperref}
\usepackage{pifont}
\usepackage{makecell}
\usepackage{diagbox}
\usepackage{setspace}
\usepackage{float}   
\usepackage{subfigure} 
\usepackage{enumitem}
\usepackage{orcidlink}
\usepackage{multirow}
\usepackage{algorithm}

\usepackage{pifont}
\usepackage{fancyhdr}
\usepackage{lipsum} % dummy text

\begin{document} 
\bstctlcite{BSTcontrol}
\linespread{1.0}

\title{
\huge TOAST: Task-Oriented Adaptive Semantic Transmission over Dynamic Wireless Environments
}
\author{
	Sheng~Yun\orcidlink{0009-0008-1011-8519},~\IEEEmembership{Student~Member,~IEEE},
    Jianhua~Pei\orcidlink{0000-0002-4066-9230},~\IEEEmembership{Student~Member,~IEEE},
	and~Ping~Wang\orcidlink{0000-0002-1599-5480},~\IEEEmembership{Fellow,~IEEE} %

	% \thanks{
	% 	Manuscript received ; revised  ; accepted . This work was supported in part by the Natural Sciences and Engineering Research Council of Canada (NSERC) Discovery Grant funded by NSERC. The associate editor coordinating the review of this article and approving it for publication was . 
	% 	\emph{(Corresponding author: Ping Wang)}
	% }%

	\thanks{
		S. Yun and P. Wang are with the Department of Electrical Engineering and Computer Science, Lassonde School of Engineering, York University, Toronto, ON, Canada (e-mails: ys97@yorku.ca; pingw@yorku.ca).
	}%

    \thanks{
 J. Pei is with the School of Electrical and Electronic Engineering, Huazhong University of Science and Technology, Wuhan, China (e-mail: jianhuapei@hust.edu.cn).
    }%

}

%\markboth{IEEE Transactions on Network Science and Engineering,~Vol.~x, No.~x, January~2026}%

\maketitle

\thispagestyle{fancy} 

\begin{abstract}
The evolution toward 6G networks demands a fundamental shift from bit-centric transmission to semantic-aware communication that emphasizes task-relevant information. This work introduces TOAST (Task-Oriented Adaptive Semantic Transmission), a unified framework designed to address the core challenge of multi-task optimization in dynamic wireless environments through three complementary components. First, we formulate adaptive task balancing as a Markov decision process, employing deep reinforcement learning to dynamically adjust the trade-off between image reconstruction fidelity and semantic classification accuracy based on real-time channel conditions. Second, we integrate module-specific Low-Rank Adaptation (LoRA) mechanisms throughout our Swin Transformer-based joint source-channel coding architecture, enabling parameter-efficient fine-tuning that dramatically reduces adaptation overhead while maintaining full performance across diverse channel impairments including Additive White Gaussian Noise (AWGN), fading, phase noise, and impulse interference. Third, we incorporate an Elucidating diffusion model that operates in the latent space to restore features corrupted by channel noises, providing substantial quality improvements compared to baseline approaches. Extensive experiments across multiple datasets demonstrate that TOAST achieves superior performance compared to baseline approaches, with significant improvements in both classification accuracy and reconstruction quality at low Signal-to-Noise Ratio (SNR) conditions while maintaining robust performance across all tested scenarios. By seamlessly orchestrating reinforcement learning, diffusion-based enhancement, and parameter-efficient adaptation within a single coherent framework, TOAST represents a significant advancement toward adaptive semantic communication systems capable of thriving in the rigorous conditions of next-generation wireless networks.

\end{abstract}


\section{Introduction}
\label{sec:intro}

\IEEEPARstart{T}{he} emergence of sixth-generation (6G) wireless networks marks a fundamental change in how communication is understood, shifting from Shannon's classical model of reliable bit transmission to a semantic-oriented approach that focuses on meaning and task relevance \cite{gunduz2023beyond}. This development in Semantic Communication (SemCom) acknowledges that, in many practical scenarios, reconstructing every bit perfectly is neither required nor efficient. Instead, the key is to retain the information necessary for completing specific tasks, such as interpreting a scene, making a decision, or initiating an action \cite{strinati2021}. This paradigm shift is particularly important as mobile applications continue to generate large volumes of visual data. Examples include autonomous vehicles that depend on real-time scene analysis and augmented reality systems that require rapid perception of the surrounding environment, all within the constraints of limited available bandwidth and energy resources.

Within this SemCom paradigm, Task-Oriented Semantic Communication (TOSC) emerges as a powerful framework that explicitly optimizes transmission for specific downstream objectives \cite{seo2023learning}. Rather than treating all information equally, TOSC systems intelligently allocate communication resources based on task relevance, transmitting only the semantic features necessary for successful task completion. This approach offers substantial efficiency gains, especially for visual communications where traditional pixel-perfect reconstruction often wastes precious bandwidth on perceptually irrelevant details. Since 2020, TOSC systems have demonstrated remarkable success across diverse applications, from object detection and scene understanding to image captioning and visual question answering \cite{qin2021semantic}. Deep Joint Source–Channel Coding (JSCC) enables end-to-end mapping of source data to channel symbols, jointly optimizing compression and protection in a task-aware manner; early convolutional models demonstrated graceful performance degradation under varying channel conditions \cite{bourtsoulatze2019deep,kurka2019deepjsccf}, and Transformer-based variants like SwinJSCC leverage hierarchical attention for superior dependency modeling \cite{yang2023swinjscc}. Diffusion-based enhancement techniques have improved perceptual quality under severe channel noise \cite{yang2024diffusion}, while vector-quantization solutions such as VQ-DeepSC boost reliability through efficient coding \cite{fu2023vq}.

Nonetheless, despite recent progress, existing TOSC systems still encounter key limitations that hinder their practical use in dynamic real-world environments. First, most current methods rely on fixed weighting schemes that fail to adapt to time-varying channel conditions or content characteristics. This poses a major challenge, as selecting the most relevant task often depends on the real-time Signal-to-Noise Ratio (SNR) and the semantic complexity of the input. Second, many systems require costly retraining or maintain multiple specialized models for different channel types, resulting in high computational and storage demands. Third, although recent studies have explored semantic importance weighting \cite{sun2023semantic}, feature-aware designs \cite{wangfeature}, and scene graph methods \cite{peiscenegraph}, they still lack a unified approach that can jointly manage multiple adaptive components such as task selection, channel adaptation, and quality enhancement within a single framework.

To overcome these limitations, we introduce TOAST: \textbf{T}ask-\textbf{O}riented \textbf{A}daptive \textbf{S}emantic \textbf{T}ransmission, a unified framework driven by three insights: \textbf{(i)} task weighting must adapt to channel conditions and content characteristics via an intelligent controller; \textbf{(ii)} efficient channel specialization relies on parameter-efficient mechanisms that preserve core model capabilities; and \textbf{(iii)} recovering perceptual quality and semantic content under severe noise requires sophisticated generative refinement. TOAST realizes these through a Swin Transformer JSCC backbone, enhanced with three core components: (1) a deep Q-network agent that continuously tunes pixel-level reconstruction versus semantic preservation weights based on channel quality, reconstruction fidelity, and classification accuracy; (2) module-specific Low-Rank Adaptation (LoRA) modules integrated into the encoder, decoder, and diffusion denoiser for rapid adaptation to Additive White Gaussian Noise (AWGN), fading, phase noise, and impulse interference with over 45× fewer parameters; and (3) an Elucidating Diffusion Model (EDM) \cite{NEURIPS2022_a98846e9} operating in latent space to restore noise-corrupted features, delivering substantial quality improvements in low-SNR regimes.

Our contributions are threefold:
\begin{itemize}
  \item We formulate loss-weight scheduling as a Markov Decision Process and employ a deep Q-network to adaptively tune weights that balance reconstruction and classification objectives, convergence, and multi-task performance across diverse channel conditions.
  \item We introduce LoRA modules into the Swin JSCC encoder, Swin decoder, and diffusion score network, enabling fast, lightweight specialization to diverse channel conditions using only a small subset of the full model parameters.
  \item We unite an end-to-end JSCC autoencoder, diffusion-based generative refinement, RL (Reinforcement Learning)-driven task prioritization, and LoRA-based adaptation into TOAST, a cohesive semantic communication framework tailored for dynamic wireless environments.
\end{itemize}
While our experiments focus on image transmission, the TOAST framework can be readily extended to other data modalities and downstream tasks with comparable benefits.

The rest of this paper is organized as follows: Section~\ref{sec:PFRW} presents the problem formulation and comprehensive related works. Section~\ref{sec:PTOAST} presents the preliminaries, including the Swin Transformer and Elucidating Diffusion Models, and further details our proposed semantic communication system architecture. Section~\ref{sec:RLATB} describes the reinforcement learning strategy for adaptive task balancing. Section~\ref{sec:CSLRA} presents the module-specific LoRA methodology. Section~\ref{sec:ERCA} provides extensive experimental results and comparative analysis. Finally, Section~\ref{sec:DLC} concludes the paper with a discussions of limitations and future directions.

\section{Problem Formulation and Related Works}
\label{sec:PFRW}

\subsection{Problem Formulation}

\subsubsection{Task-Oriented Semantic Communication System}

We consider a task-orineted SemCom system designed to transmit images over wireless channels while preserving both visual fidelity and semantic content required for downstream tasks. In this work, the downstream task is specifically formulated as a multi-class classification problem. Let $\mathbf{x} \in \mathbb{R}^{H \times W \times C}$ denote an input image with height $H$, width $W$, and $C$ channels, and let $y \in \mathcal{Y}$ represent the corresponding semantic label, where the label space is defined as $\mathcal{Y} = \{1, 2, \ldots, K\}$ for a $K$-class classification task.

The task-oriented SemCom system consists of three primary components:

\begin{itemize}
    \item \textbf{Transmitter:} An encoder $f_{\text{enc}}: \mathbb{R}^{H \times W \times C} \rightarrow \mathbb{R}^{L}$ that maps the input image $\mathbf{x}$ to a compact latent representation $\mathbf{z} \in \mathbb{R}^{L}$, where $L \ll H\times W \times C$ denotes the compressed dimensionality. This encoding process aims to preserve both visual and semantic information in a bandwidth-efficient form.

    \item \textbf{Channel:} A wireless channel $\mathcal{C}: \mathbb{R}^{L} \rightarrow \mathbb{R}^{L}$ that simulates real-world transmission impairments by introducing noise and distortion. The received signal is denoted by $\mathbf{z}_{\text{ch}} = \mathcal{C}(\mathbf{z})$.

    \item \textbf{Receiver:} A dual-head architecture that simultaneously performs image reconstruction and semantic inference. It comprises: \textbf{(i)} a reconstruction decoder $f_{\text{dec}}: \mathbb{R}^{L} \rightarrow \mathbb{R}^{H \times W \times C}$, which generates the reconstructed image $\hat{\mathbf{x}} = f_{\text{dec}}(\mathbf{z}_{\text{ch}})$; and \textbf{(ii)} a semantic classifier $f_{\text{cls}}: \mathbb{R}^{L} \rightarrow \mathbb{R}^{K}$, which outputs class logits $\hat{\mathbf{y}} = f_{\text{cls}}(\mathbf{z}_{\text{ch}})$ corresponding to the $K$ semantic classes.
\end{itemize}

\subsubsection{Channel Model}

We model the wireless channel as an additive noise process with channel-specific characteristics. For the general case, the received signal is given by:
\begin{equation}
\mathbf{z}_{\text{ch}} = \mathbf{h} \odot \mathbf{z}_{\text{norm}} + \mathbf{n},
\end{equation}
where $\mathbf{z}_{\text{norm}}$ is the power-normalized transmitted signal satisfying $\mathbf{z}_{\text{norm}} = \mathbf{z}/\sqrt{\mathbb{E}[\|\mathbf{z}\|^2]}$, $\mathbf{h}$ represents the channel coefficient (which may be deterministic for AWGN or random for fading channels), $\odot$ denotes element-wise multiplication, and $\mathbf{n}$ represents additive noise. The signal-to-noise ratio (SNR) is defined as $\text{SNR} = \mathbb{E}[\|\mathbf{z}_{\text{norm}}\|^2] / \mathbb{E}[\|\mathbf{n}\|^2]$.

For our experiments, we consider both additive noise and fading scenarios, including AWGN (where $\mathbf{h} = \mathbf{1}$), Rayleigh fading (where $\mathbf{h}$ follows a complex Gaussian distribution), Rician fading (incorporating line-of-sight components), phase noise (introducing phase distortions), and impulse noise (modeling bursty interference).

\subsubsection{Multi-Task Optimization Objective}

The task-oriented SemCom system addresses a dual-objective optimization problem that balances reconstruction quality and semantic preservation:
\begin{equation}
\min_{\theta} \mathbb{E}_{\mathbf{x}, y, \mathcal{C}} \left[ \lambda_{\text{recon}} \mathcal{L}_{\text{recon}}(\mathbf{x}, \hat{\mathbf{x}}) + \lambda_{\text{cls}} \mathcal{L}_{\text{cls}}(y, \hat{\mathbf{y}}) \right],
\end{equation}
where $\mathcal{L}_{\text{recon}}$ measures reconstruction fidelity (e.g., Mean Squared Error (MSE) or perceptual/semantic loss), $\mathcal{L}_{\text{cls}}$ measures classification performance (e.g., cross-entropy loss), $\lambda_{\text{recon}}, \lambda_{\text{cls}} \geq 0$ are task-specific weighting parameters satisfying $\lambda_{\text{recon}} + \lambda_{\text{cls}} = 1$, and $\theta$ encompasses all trainable parameters in the system.

\subsubsection{Key Challenges and Design Requirements}

This formulation reveals several critical challenges that motivate the proposed approach:

\begin{itemize}
    \item \textbf{Multi-Task Trade-off}: The reconstruction and classification objectives often compete for limited representational capacity in the latent space, requiring intelligent balancing strategies.
    
    \item \textbf{Channel Adaptivity}: Optimal task weighting ($\lambda_{\text{recon}}$, $\lambda_{\text{cls}}$) depends dynamically on channel conditions—lower SNRs typically favor reconstruction preservation while higher SNRs enable greater emphasis on semantic tasks.
    
    \item \textbf{Performance Stability}: The system must maintain acceptable performance across a wide range of SNR conditions without requiring channel state information at the transmitter.
    
    \item \textbf{Computational Efficiency}: Adaptation to varying channel conditions should not require full model retraining, necessitating parameter-efficient adaptation mechanisms.
    
    \item \textbf{Content Awareness}: Different image types and semantic complexities may require different balancing strategies, demanding content-adaptive optimization.
\end{itemize}

These challenges collectively motivate the need for an adaptive, multi-objective optimization framework that can dynamically balance competing tasks while efficiently adapting to diverse operating conditions. Our proposed TOAST framework addresses these requirements through a combination of reinforcement learning-based task balancing, diffusion-enhanced denoising, and parameter-efficient adaptation mechanisms.

\subsection{Related Works}

SemCom fundamentally departs from Shannon's information theory by prioritizing meaning extraction over reliable bit transmission \cite{gunduz2023beyond, strinati2021}, focusing on task-relevant information preservation for efficient resource utilization \cite{seo2023learning}. This paradigm shift, enabled by advances in deep learning and foundation models \cite{10183789}, has positioned SemCom as essential for 6G applications, including extended reality and autonomous systems \cite{strinati2021}. Early deep JSCC models demonstrate that convolutional autoencoders could map images directly to channel symbols with graceful SNR degradation \cite{bourtsoulatze2019deep, kurka2019deepjsccf}, while Transformer-based variants like SwinJSCC leverage hierarchical self-attention for improved dependency modeling \cite{yang2023swinjscc}. However, these approaches remain anchored to reconstruction metrics, relegating semantic tasks to post-processing rather than end-to-end optimization.

\begin{table*}[t!]
\centering
\caption{Comparison of Recent Semantic Communication Frameworks (2020-2025)}
\label{tab:comparison}
\footnotesize
\begin{tabular}{|p{3.9cm}|c|c|c|c|>{\centering\arraybackslash}p{4.5cm}|}
\hline
\textbf{Framework} & \textbf{End-to-End} & \textbf{Channel} & \textbf{Parameter} & \textbf{Quality} & \textbf{Tasks Supported} \\
\textbf{(Reference)} & \textbf{Learning} & \textbf{Adaptivity} & \textbf{Efficiency} & \textbf{Enhancement} & \\
\hline
\hline
Lyu et al. \cite{lyu2024multi} & Yes & Yes & No & No & Image reconstruction, classification \\
\hline
Sun et al. (GRACE) \cite{sun2024grace} & Yes & Yes & No & No & Image retrieval \\
\hline

Wu et al. (CDDM) \cite{wu2024cddm}              & Yes & Yes & No & Yes & Image reconstruction\\ 
\hline
Yang et al. (Diff-JSCC) \cite{yang2024diffusion} & Yes & No & No & Yes & Image reconstruction\\ 
\hline
Niu et al. (Hybrid-Diff) \cite{niu2023hybrid}    & Yes & Yes & No & Yes & Image reconstruction\\ 
\hline
Eldeeb et al. \cite{eldeeb2024cav} & Yes & No & No & No & Road sign reconstruction, classification \\
\hline
Ma et al. ($\beta$-VAE) \cite{ma2023explainable} & No & Yes & No & No & Classification (explainable) \\
\hline
Wu et al. (STAT) \cite{wu2022stat} & No & No & Yes & No & Classification, object detection \\
\hline
Park et al. \cite{park2024joint} & Yes & Yes & No & No & Classification, reconstruction, retrieval \\
\hline
Wang et al. (CA-DJSCC) \cite{wang2025ca} & Yes & Yes & No & No & Image reconstruction, classification \\
\hline
Seon et al. \cite{seon2024dual} & No & Yes & No & No & Image classification \\
\hline
Wang et al. (I-JSCC) \cite{wangfeature} & Yes & Yes & No & No & Image classification \\
\hline
Liu et al. (ASC) \cite{liu2023adaptable} & No & Yes & No & No & Generic task-oriented inference \\
\hline
Huang et al. (DSSCC) \cite{huang2024dsscc} & Yes & No & No & No & Image reconstruction, classification \\
\hline
He et al. (DeepSC-MM) \cite{he2023rate} & Yes & Yes & No & No & VQA, sentiment analysis \\
\hline
Fu et al. (DA-TOSC) \cite{fu2024hybrid} & Yes & Yes & Yes & No & Various AI tasks \\
\hline
Shao et al. (IB) \cite{shao2022ib} & Yes & Yes & No & No & Image classification \\
\hline
Huang et al. \cite{huang2023dl} & Yes & No & No & No & Image transmission \\
\hline
Xie et al. (Robust IB) \cite{xie2023robust} & Yes & Yes & No & No & Image classification \\
\hline
Gao et al. \cite{gao2024adaptive} & No & Yes & No & No & Generic semantic tasks \\
\hline
Yang et al. (WITT) \cite{yang2023witt} & Yes & No & No & No & Image reconstruction \\
\hline
Wu et al. (OFDM) \cite{wu2022ofdm} & Yes & Yes & No & Yes & Image reconstruction \\
\hline
Yang \& Kim (OFDM) \cite{yang2022ofdm} & Yes & Yes & No & No & Image reconstruction \\
\hline
Yang \& Kim (Rate) \cite{yang2022rate} & Yes & Yes & No & No & Image reconstruction \\
\hline
Zhang et al. \cite{zhang2023predictive} & Yes & Yes & No & No & Image reconstruction \\
\hline
Fu et al. (VQ-DeepSC) \cite{fu2023vq} & Yes & No & No & No & Image reconstruction \\
\hline
Hu et al. (Masked VQ) \cite{hu2023masked} & Yes & Yes & No & No & Image transmission \\
\hline
Song et al. \cite{song2020adversarial} & Yes & Yes & No & No & Image reconstruction \\
\hline
Tung et al. \cite{tung2022deepjsccq} & Yes & No & No & No & Image transmission \\
\hline
Bo et al. \cite{bo2022joint} & Yes & No & No & No & Semantic data delivery \\
\hline
Zhang et al. (Multi-level) \cite{zhang2025multi} & Yes & Yes & Yes & No & Classification, detection, reconstruction \\
\hline
Chen et al. \cite{chen2023personalized} & Yes & No & No & No & Personalized image inference \\
\hline

Sagduyu et al. \cite{sagduyu2023task} & Yes & No & No & No & Classification \\
\hline
Zhang et al. \cite{9953099} & Yes & No & No & No & Classification \\
\hline
Qiao et al. \cite{qiao2024latency} & Yes & No & No & Yes & Image reconstruction \\
\hline
Pei et al. \cite{10896580} & Yes & Yes & No & Yes & Image reconstruction, semantic disambiguation \\
\hline
Sheng et al. \cite{sheng2022multi} & Yes & No & No & No & Text classification, translation, summarization \\
\hline
Wang et al. \cite{wang2024unified} & Yes & No & Yes & No & Multiple NLP and vision tasks
 \\
\hline

\textbf{TOAST (Ours)} & \textbf{Yes} & \textbf{Yes} & \textbf{Yes} & \textbf{Yes} & \begin{tabular}[c]{@{}c@{}}\textbf{Image reconstruction, classification,}\\\textbf{and other various AI tasks}\end{tabular}  \\
\hline
\end{tabular}

\end{table*}

Recent advances have addressed these limitations through several innovative directions, as comprehensively analyzed in our survey of 38 representative frameworks (Table \ref{tab:comparison}): 
\begin{itemize}
\item In \textbf{task‐aware improvements}, semantic importance weighted JSCC reweights features based on classification relevance \cite{sun2023semantic}, while feature importance aware frameworks systematically prioritize task‐critical information \cite{wangfeature}. Deep Neural Networks (DNNs) \cite{sagduyu2023task} and Generative Adversarial Networks (GANs) \cite{9953099} based JSCC approaches improve semantic quality and classification performance by incorporating classification tasks at the receiver. Scene graph approaches enable selective transmission of task‐relevant elements with frameworks like GRACE \cite{sun2024grace} introducing semantic-aware adaptive channel coding that masks less important scene-graph features according to their significance and current channel state. Multi-task JSCC systems have emerged that simultaneously support image reconstruction and classification \cite{lyu2024multi,eldeeb2024cav,wang2025ca}, achieving up to 89\% bandwidth savings in autonomous vehicle applications \cite{eldeeb2024cav}. Notably, explainable frameworks based on $\beta$-VAE (Variational Autoencoder) \cite{ma2023explainable} disentangle latent features to transmit only task-relevant semantic components, addressing the black-box nature of traditional approaches.
\item To \textbf{mitigate wireless channel noises and distortions}, generative methods such as Channel Denoising Diffusion Models (CDDM) iteratively refine corrupted embeddings \cite{wu2024cddm}, latency aware frameworks synthesize missing details under stringent delay constraints \cite{qiao2024latency}, and hybrid separation diffusion schemes achieve competitive accuracy at reduced bitrates \cite{niu2023hybrid}. Based on the CDDM denoising mechanism, the application of knowledge distillation and adaptor can significantly improve the efficiency and adaptability of channel denoising \cite{10896580}. Nonetheless, these channel denoising diffusion approaches have not considered the presence of multiple tasks at the receiving end. Our analysis reveals that 23 out of 38 surveyed frameworks now incorporate channel-adaptive mechanisms, ranging from gating networks that prune features based on SNR \cite{lyu2024multi} to multi-agent Deep Reinforcement Learning (DRL) systems that adapt both transmission and reception strategies without explicit Channel State Information (CSI) \cite{seon2024dual}. Particularly innovative are reference signal-based implicit CSI estimation approaches \cite{wang2025ca} that achieve over 40\% reduction in channel usage, and predictive coding schemes \cite{zhang2023predictive} that anticipate channel variations through feedback loops. Robust training strategies have also evolved, including adversarial bit-flip training \cite{song2020adversarial}, masked codeword learning \cite{hu2023masked}, and multi-SNR optimization \cite{xie2023robust}, enabling systems to maintain performance across diverse channel conditions without external enhancement modules.
\item For \textbf{adaptation in dynamic wireless environments and fine‐tuning of pre‐trained large visual models}, multi‐task SemCom systems have explored shared encoders with task‐specific decoders \cite{sheng2022multi} and hierarchical knowledge bases \cite{wang2024unified}, while channel adaptive mechanisms adjust compression levels according to channel conditions \cite{liu2023adaptable,park2024joint,he2023rate}. Transfer learning approaches like STAT \cite{wu2022stat} enable knowledge sharing between classification and detection tasks with minimal fine-tuning, while multi-modal frameworks \cite{he2023rate} support complex tasks like VQA through unequal error protection across modalities. The hybrid digital-analog framework \cite{fu2024hybrid} addresses compatibility challenges by combining analog semantic features with digital reliability, enabling deployment within existing infrastructures. Moreover, LoRA has emerged as a parameter efficient fine‐tuning technique, significantly reducing trainable parameters while preserving performance \cite{hu2021lora}, though our survey reveals only 4 frameworks—STAT \cite{wu2022stat}, DA-TOSC \cite{fu2024hybrid}, Multi-Level \cite{zhang2025multi}
and \cite{wang2024unified}—currently exploit parameter-efficient adaptation-related approaches.
\end{itemize}

Despite these advances, existing TOSC systems face several limitations that directly motivate our problem formulation as follows:
\begin{itemize}
\item First, most approaches rely on static weighting between reconstruction and semantic objectives throughout operation, which is a critical gap given our formulation's emphasis on the multi-task trade-off challenge. This fails to capture the dynamic nature of wireless channels, where optimal trade-offs between fidelity and semantics vary with uncertain SNR and content complexity. As shown in Table \ref{tab:comparison}, while 33 frameworks employ end-to-end learning, they typically fix the task balancing at training time, unable to adapt to the channel-dependent optimal weighting our formulation identifies as crucial. Fixed weighting schemes struggle to adapt to low-SNR scenarios that require prioritizing reconstruction or to high-SNR regimes that allow greater emphasis on semantic discrimination.
\item Second, current adaptation methods for SemCom systems typically incur high computational costs \cite{10896580}, often requiring full or partial retraining when channel conditions change, which directly conflicts with our requirement for computational efficiency. Our survey reveals that despite various channel-adaptive mechanisms, only 4 out of 38 frameworks incorporate parameter-efficient adaptation techniques. Most multi-task systems still require separate models or substantial retraining for new tasks, failing to address the parameter efficiency challenge identified in our problem formulation.
\item Third, the integration of multiple adaptive mechanisms such as dynamic loss scheduling, generative refinement, and parameter-efficient tuning remains unexplored within a unified framework. While some works achieve channel adaptivity \cite{lyu2024multi,sun2024grace,wang2025ca} and others focus on multi-task support \cite{eldeeb2024cav,he2023rate,zhang2025multi}, none combine these with both parameter-efficient adaptation and quality enhancement in a single system. This fragmentation prevents existing systems from effectively addressing all the key challenges outlined in our problem formulation, including multi-task trade-offs, channel adaptivity, performance stability, computational efficiency, and content awareness.
\end{itemize}

To address these gaps, we propose TOAST, a task-oriented adaptive SemCom framework that uniquely addresses all five challenges identified in our problem formulation. TOAST formulates the joint multi-task objective as a Markov decision process, where a reinforcement learning agent observes channel conditions, performance metrics, and training progress to continuously adjust task weightings, enabling on-the-fly adaptation without manual intervention. The framework integrates diffusion-enhanced reconstruction to improve robustness under severe distortion and employs module-specific parameter-efficient adaptation (e.g., LoRA) to rapidly recalibrate model parameters with minimal overhead. By unifying dynamic task balancing and generative refinement in a single pipeline, TOAST achieves superior performance across multiple objectives in fluctuating wireless environments. While demonstrated on image reconstruction and classification tasks, the framework's modular design and adaptive weighting mechanism can be easily extended to other task combinations, such as detection, segmentation, or multi-modal tasks, with minimal architectural changes. This makes it the first framework to comprehensively address all aspects of our multi-objective optimization problem while retaining the flexibility to support a wide range of semantic communication applications.

\section{Proposed Task-Oriented Adaptive Semantic Transmission Framework}
\label{sec:PTOAST}

This section presents our comprehensive TOAST framework that addresses the challenges identified in Section~\ref{sec:PFRW} through an integration of multiple adaptive components. We begin by introducing the foundational architectures that form the backbone of our system—the Swin Transformer for joint source-channel coding and the EDM for enhanced reconstruction. We then detail how these components are orchestrated within our unified framework, incorporating reinforcement learning-based task balancing and parameter-efficient channel adaptation.

\begin{figure*}[t!]
  \centering
  \includegraphics[width=1\textwidth]{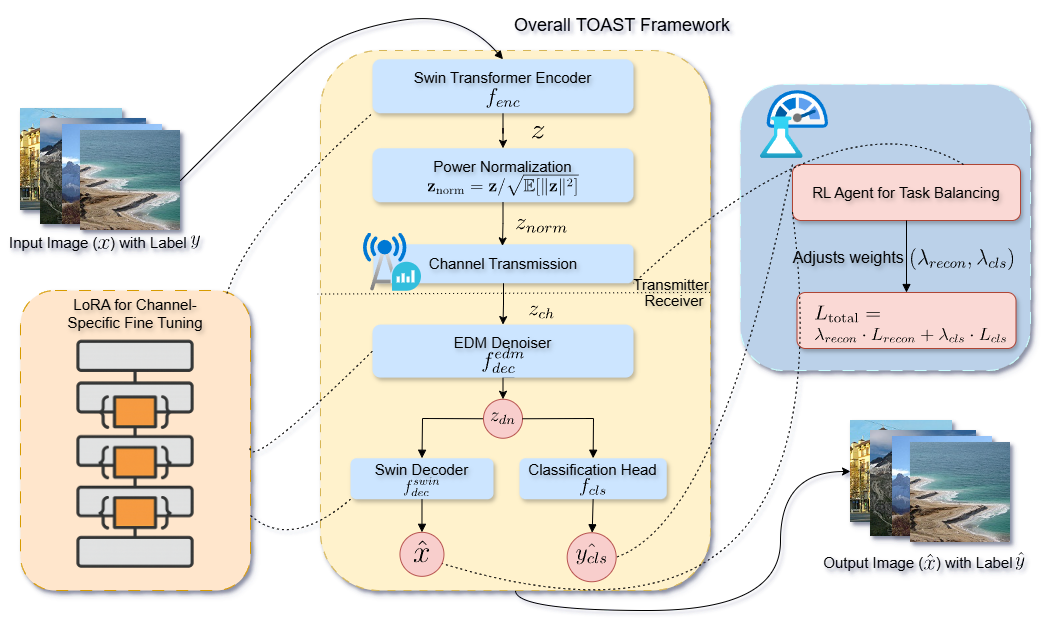}
  \caption{System overview of the proposed semantic communication framework. The input image is encoded via a Swin Transformer to produce a latent code transmitted over a noisy channel. The received latent is first denoised using an Elucidating Diffusion Model (EDM), then passed to both the Swin decoder for image reconstruction and a classification head for semantic prediction. Reinforcement Learning dynamically adjusts task priorities by tuning the reconstruction and classification loss weights based on channel feedback. LoRA modules enable efficient, channel-specific adaptation across model components. Algorithm 1 variable $\mathbf{z}_{\text{ch}}$ appears in italic font within the diagram.}
  \label{fig:system_overview}
\end{figure*}

\subsection{Swin Transformer for Joint Source-Channel Coding}

The Swin Transformer serves as the core architecture of our JSCC scheme for wireless communication. By leveraging a hierarchical attention mechanism with shifted windows, it captures both local and global dependencies while maintaining linear computational complexity with respect to image size. The following three components summarize its design:

\subsubsection{Hierarchical Feature Extraction via Windowed Attention}
Let an input image be denoted by $\mathbf{x}\in\mathbb{R}^{H\times W\times C}$. In the encoder, $\mathbf{x}$ is first partitioned into non-overlapping patches of size $P\times P$. Each patch is linearly embedded into a $D$-dimensional feature vector through a patch embedding layer, yielding a sequence suitable for Transformer processing. Attention is computed within local windows of size $M\times M$. For queries $\mathbf{Q}$, keys $\mathbf{K}$, and values $\mathbf{V}$, the self-attention operation in each window is:
\begin{equation}
   \text{Attention}(\mathbf{Q}, \mathbf{K}, \mathbf{V}) = \text{Softmax}\left(\frac{\mathbf{Q}\mathbf{K}^T}{\sqrt{d_k}} + \mathbf{B}\right)\mathbf{V},
\end{equation}
where $\mathbf{B}$ denotes learnable relative position biases that encode spatial relationships. To enable cross-window connections without incurring quadratic complexity, the shifted window mechanism displaces the partition grid by $(\lfloor M/2\rfloor,\lfloor M/2\rfloor)$ pixels in alternating layers. Consequently, complexity is reduced from $O((HW)^2)$ to $O(M^2HW)$ while preserving modeling capacity.

\subsubsection{Hierarchical Feature Aggregation and Channel Coding}
As the network deepens, patch merging layers progressively reduce spatial resolution and increase channel dimensionality. Each merging step concatenates features from a $2\times 2$ block of neighboring patches and applies a linear projection to control dimensionality growth. After several stages of alternating windowed attention and patch merging, the hierarchical features are flattened and projected to form a latent code $\mathbf{z}\in\mathbb{R}^L$. This latent vector is then power-normalized according to the wireless channel constraints described in Section~\ref{sec:PFRW}.

\subsubsection{Hierarchical Decoding and Reconstruction}
The decoder mirrors the encoder’s structure in reverse. Patch expanding layers are used to upsample features at each stage: features from a lower resolution are linearly projected and reshaped into a $2\times 2$ spatial block, restoring higher resolution while preserving learned representations. Windowed attention (using the same shifted mechanism) is applied at each upsampled stage to maintain spatial coherence. Finally, a linear projection recovers an image $\hat{\mathbf{x}}\in\mathbb{R}^{H\times W\times C}$ from the highest-resolution feature map. End-to-end training is accomplished by minimizing a reconstruction loss (e.g., MSE or perceptual/semantic metrics) under the JSCC objective.

\subsection{Elucidating Diffusion Model for Latent Space Denoising}

While the Swin Transformer provides robust feature extraction and reconstruction capabilities, channel noise can significantly degrade the transmitted latent representations. To mitigate this issue, we employ the EDM \cite{NEURIPS2022_a98846e9} as a denoiser operating in the latent space. Diffusion models have demonstrated strong generative performance for semantic communication, with methods such as Denoising Diffusion Probabilistic Models (DDPM)\cite{ho2020denoisingdiffusionprobabilisticmodels}, CDDM\cite{wu2024cddm}, and diffusion-aided joint source channel coding \cite{yang2024diffusion} yielding notable improvements in perceptual quality under severe noise conditions. However, traditional diffusion approaches require hundreds of iterative denoising steps to achieve acceptable quality \cite{10542391}, which presents a critical limitation for low-latency SemCom systems~\cite{10896580}. The EDM overcomes these drawbacks through its continuous time framework and variance preserving preconditioning, enabling high quality denoising with substantially fewer steps.

\subsubsection{Continuous-Time Diffusion Framework}
EDM models the denoising task as the reversal of a continuous time diffusion process. In contrast to discrete schemes, the forward diffusion in EDM is defined by:
\begin{equation}
   d\mathbf{z}_t = -\tfrac{1}{2}\beta(t)\,\mathbf{z}_t\,dt \;+\; \sqrt{\beta(t)}\,d\mathbf{w}_t,
\end{equation}
where $\mathbf{z}_t$ is the latent representation at time $t$, $\beta(t)$ denotes the noise schedule and $\mathbf{w}_t$ represents Brownian motion. The corresponding reverse process uses a learned score network $s_\theta(\mathbf{z}_t, t)$ to approximate the score function $\nabla_{\mathbf{z}_t}\log p_t(\mathbf{z}_t)$, where $p_t$ denotes its marginal distribution under this process, yielding:
\begin{equation}
   d\mathbf{z}_t = \Bigl[\tfrac{1}{2}\beta(t)\,\mathbf{z}_t \;+\; \beta(t)\,s_\theta(\mathbf{z}_t, t)\Bigr]\,dt.
\end{equation}

\subsubsection{Preconditioning and Sampling Efficiency}

A key innovation in EDM is its Variance-Preserving (VP) formulation with carefully designed preconditioning:
\begin{equation}
s_\theta(\mathbf{z}, \sigma) = \frac{1}{\sigma}D_\theta(c_{\text{in}}(\sigma)\mathbf{z}; c_{\text{noise}}(\sigma)),
\end{equation}
where $c_{\text{in}}(\sigma)$ and $c_{\text{noise}}(\sigma)$ are preconditioning functions that stabilize training across noise levels. This formulation enables the use of deterministic Ordinary Differential Equation (ODE) solvers (e.g., Heun's method \cite{heun1900neue}) for faster inference compared to stochastic sampling, achieving high-quality denoising.

By applying EDM in the latent space, the denoiser learns to distinguish meaningful signal variations from channel induced noise artifacts. Operating on compact latent representations rather than raw pixels yields computational efficiency while preserving semantic information essential for downstream reconstruction and classification tasks. The continuous time formulation and variance preserving preconditioning together enable high quality denoising with minimal sampling steps, making EDM an effective component for low-latency JSCC and TOSC systems.

\subsection{Integrated Architecture and Adaptive Components}
\begin{algorithm}[t!]
\caption{TOAST: Task-Oriented Adaptive Semantic Transmission}
\label{alg:system_workflow}
\begin{algorithmic}[1]
\Require Input image $\mathbf{x} \in \mathbb{R}^{H \times W \times C}$, ground truth label $y$, channel $\mathcal{C}$
\Ensure Reconstructed image $\hat{\mathbf{x}}$, predicted label $\hat{\mathbf{y}}$

\State \textbf{// RL-Based Adaptive Task Weighting}
\State Observe state: $s_t \leftarrow [\text{SNR}_{\text{norm}}, L_{\text{recon}}^{\text{norm}}, \text{Acc}_{\text{cls}}, P_{\text{epoch}}, \lambda_{\text{recon}}^{\text{prev}}]$
\State Predict weights: $[\lambda_{\text{recon}}, \lambda_{\text{cls}}] \leftarrow \pi_\theta(s_t)$ with $\lambda_{\text{recon}} + \lambda_{\text{cls}} = 1$

\State \textbf{// Encoding and Channel Transmission}
\State Load channel-specific LoRA: $\Delta W_c = \alpha_c B_c A_c$ for channel type $c$
\State Encode with adaptation: $\mathbf{z} \leftarrow f_{\text{enc}}(\mathbf{x}; W + \Delta W_{\text{enc}})$
\State Normalize and transmit: $\mathbf{z}_{\text{ch}} \leftarrow \mathcal{C}(\mathbf{z}/\sqrt{\mathbb{E}[\|\mathbf{z}\|^2]})$

\State \textbf{// EDM Latent Denoising}
\State Estimate noise level: $\sigma_{\max} \leftarrow \|\mathbf{z}_{\text{ch}} - \mathbf{z}_{\text{norm}}\|_2$

\For{$t = T$ \textbf{down to} $1$}
    \State $\mathbf{z}_{\text{ch}} \leftarrow \mathbf{z}_{\text{ch}} - \sigma_t^2 \cdot s_{\theta}(\mathbf{z}_{\text{ch}}, \sigma_t; W + \Delta W_e)$
\EndFor
\State $\mathbf{z}_{\text{dn}} \leftarrow \mathbf{z}_{\text{ch}}$

\State \textbf{// Dual-Task Decoding}
\State Reconstruct: $\hat{\mathbf{x}} \leftarrow f_{\text{dec}}(\mathbf{z}_{\text{dn}}; W + \Delta W_d)$
\State Classify: $\hat{\mathbf{y}} \leftarrow f_{\text{cls}}(\mathbf{z}_{\text{dn}}; W + \Delta W_c)$

\State \textbf{// Loss Computation and Learning}
\State $\mathcal{L}_{\text{total}} \leftarrow \lambda_{\text{recon}} \mathcal{L}_{\text{recon}}(\mathbf{x}, \hat{\mathbf{x}}) + \lambda_{\text{cls}} \mathcal{L}_{\text{cls}}(y, \hat{\mathbf{y}})$
\State Compute reward: $R_t \leftarrow f_R(\Delta \mathcal{L}_{\text{recon}}, \Delta \text{Acc}_{\text{cls}})$
\State Update RL policy: $\pi_\theta \leftarrow \pi_\theta - \eta_{\text{RL}} \nabla_\theta \mathcal{L}_{\text{DQN}}$
\State Update LoRA adapters: $\{B_c, A_c\} \leftarrow \{B_c, A_c\} - \eta \nabla \mathcal{L}_{\text{total}}$

\State \Return $\hat{\mathbf{x}}, \hat{\mathbf{y}}$
\end{algorithmic}
\end{algorithm}
Building upon the Swin Transformer backbone and EDM denoiser, our complete TOAST framework integrates reinforcement learning for dynamic task loss balancing and LoRA for efficient channel-specific tuning. Fig.~\ref{fig:system_overview} illustrates the full system architecture. The core processing pipeline and its two key enhancements are detailed as follows:

\subsubsection{Core Processing Pipeline}

The end-to-end processing flow operates as follows:

\begin{itemize}
    \item \textbf{Encoding}: The Swin encoder $f_{\text{enc}}$ maps the input image through hierarchical attention blocks, producing latent representation $\mathbf{z}$.
    
    \item \textbf{Channel Transmission}: After power normalization, $\mathbf{z}_{\text{norm}}$ traverses the noisy channel, yielding $\mathbf{z}_{\text{ch}} = \mathcal{C}(\mathbf{z}_{\text{norm}})$.
    
    \item \textbf{Latent Denoising}: The EDM denoiser $f_{\text{dec}}^{\text{edm}}$ processes $\mathbf{z}_{\text{ch}}$ to produce a refined latent $\mathbf{z}_{\text{dn}}$, mitigating channel-induced corruptions.
    
    \item \textbf{Dual-Task Decoding}: The denoised latent feeds both (i) the Swin decoder $f_{\text{dec}}^{\text{swin}}$ for image reconstruction $\hat{\mathbf{x}}$, and (ii) an Multilayer Perceptron classifier $f_{\text{cls}}$ for semantic prediction $\hat{\mathbf{y}}$.
\end{itemize}

\subsubsection{Reinforcement Learning for Dynamic Task Balancing}

To address the challenge of SNR-dependent optimal weighting identified in Section~\ref{sec:PFRW}, we propose a reinforcement learning approach that automatically adapts task weights based on real-time channel conditions. We formulate the task balancing problem as a Markov Decision Process where an RL agent continuously observes the system state and dynamically adjusts the weighting parameters $\lambda_{\text{recon}}$ and $\lambda_{\text{cls}}$ without requiring manual hyperparameter tuning.

The RL agent takes as input a comprehensive state representation that includes normalized SNR conditions, current reconstruction loss, classification accuracy, training progress, and previous weight settings. Based on this state information, the agent outputs weight adjustments that optimize both immediate performance improvements and long-term adaptation to varying channel conditions. The reward function is designed to encourage balanced improvements in both reconstruction quality and classification accuracy while promoting exploration of the weight space to discover optimal configurations across different SNR regimes.

This adaptive approach eliminates the need for extensive hyperparameter searches and enables the system to automatically respond to changing wireless conditions during deployment. The detailed design methodology is discussed in Section \ref{sec:RLATB}.

\subsubsection{Low-Rank Adaptation for Channel-Specific Tuning}

Recognizing that different channel types (e.g., AWGN, Rayleigh fading, phase noise, etc.) require distinct processing strategies, we inject LoRA modules throughout the architecture. For each adapted weight matrix:
\begin{equation}
W' = W + \alpha_c B_c A_c, \quad B_c \in \mathbb{R}^{d \times r_c}, \quad A_c \in \mathbb{R}^{r_c \times k},
\end{equation}
where $r_c \ll \min(d,k)$ is the module-specific rank, $W$ denotes the original pretrained weight matrix, $W'$ the adapted weight after LoRA update, and $\alpha_c$ a channel-specific scaling factor that modulates the influence of the low-rank adjustment. During channel-specific fine-tuning, only the low-rank matrices $(B_c, A_c)$ are updated, preserving base model performance while enabling rapid adaptation with minimal computational overhead. Section \ref{sec:CSLRA} presenters an in-depth exposition of the proposed method.

\subsubsection{Algorithmic Overview}

Algorithm~\ref{alg:system_workflow} details the complete inference workflow, illustrating how these components interact to process images through our TOAST Framework. This integrated architecture embodies our vision of an adaptive TOSC system capable of robustly operating under diverse channel conditions while preserving both reconstruction fidelity and semantic accuracy. The coordinated design of hierarchical attention, diffusion-based denoising, reinforcement-driven task balancing, and parameter-efficient adaptation ensures reliable performance across the demanding scenarios of next-generation mobile wireless networks.

\section{Reinforcement Learning for Adaptive Task Balancing}
\label{sec:RLATB}

Motivated by the limitations of static weighting schemes identified in Section~\ref{sec:PFRW}, this section presents our reinforcement learning framework for adaptive task balancing. We formulate the dynamic weight adjustment problem as a Markov Decision Process (MDP), enabling an intelligent agent to continuously optimize the trade-off between reconstruction fidelity and semantic preservation based on real-time channel conditions and performance feedback.

\subsection{RL Framework Design}

We formulate the adaptive task balancing problem as a MDP, where an intelligent agent learns to dynamically adjust the relative importance of reconstruction and classification objectives based on observed system states. This formulation naturally captures the sequential nature of the optimization process and enables the system to discover complex adaptation strategies through experience. 

\subsubsection{State Space Formulation}

The design of an effective state representation is crucial for enabling the RL agent to make informed decisions. Our state space captures both the current operating conditions and recent performance history:
\begin{equation}
s_t = [\text{SNR}_{\text{norm}}, L_{\text{recon}}^{\text{norm}}, \text{Acc}_{\text{cls}}, P_{\text{epoch}}, \lambda_{\text{recon}}^{\text{prev}}],
\end{equation}
where each component provides essential context:
\begin{itemize}
    \item $\text{SNR}_{\text{norm}} \in [0, 1]$: Normalized channel SNR computed as $\text{SNR}_{\text{current}}/\text{SNR}_\text{max}$, where the $\text{SNR}_{\text{current}}$ 
 and $\text{SNR}_\text{max}$ denote the present and the maximum SNR, respectively, providing direct insight into transmission reliability;
    \item $L_{\text{recon}}^{\text{norm}} \in [0, 1]$: Normalized reconstruction loss using exponential moving average: $L_{\text{recon}}^{\text{norm}} = L_{\text{recon}}^{\text{current}}/L_{\text{recon}}^{\text{EMA}}$, where $L_{\text{recon}}^{\text{current}}$ is the loss on the latest batch and $L_{\text{recon}}^{\text{EMA}}$ its exponential moving average, providing a stable fidelity baseline;
    \item $\text{Acc}_{\text{cls}} \in [0, 1]$: Classification accuracy on recent batches (naturally normalized);
    \item $P_{\text{epoch}} \in [0, 1]$: Normalized training progress calculated as the proportion of completed training epochs to offer the RL agent phase-aware adaptation;
    \item $\lambda_{\text{recon}}^{\text{prev}} \in [0, 1]$: Previous reconstruction weight from the last time step, providing temporal context;
\end{itemize}

This state representation enables the agent to recognize channel characteristics and performance trends, facilitating intelligent weight adjustment decisions. The exponential moving average for loss normalization ensures stability against training dynamics, while SNR clipping to the operational range prevents outlier effects.

\subsubsection{Action Space Formulation}

The agent's actions directly control the task weighting in our multi-objective loss function:

\begin{equation}
a_t = [\lambda_{\text{recon}}, \lambda_{\text{cls}}], \quad \text{subject to } \lambda_{\text{recon}} + \lambda_{\text{cls}} = 1.
\end{equation}
These weights modulate the composite training objective:
\begin{equation} \label{eq:multiloss}
\mathcal{L}_{\text{total}} = \lambda_{\text{recon}} \cdot \mathcal{L}_{\text{recon}}(\mathbf{x}, \hat{\mathbf{x}}) + \lambda_{\text{cls}} \cdot \mathcal{L}_{\text{cls}}(y, \hat{\mathbf{y}}).
\end{equation}
The constraint ensures consistent loss scaling while providing the flexibility to shift emphasis between objectives based on current conditions.

\subsubsection{Reward Function Design}

The design of an effective reward function is crucial for training an RL agent that can balance competing objectives while maintaining exploratory behavior. Our reward function must incentivize immediate performance improvements on both tasks without causing the agent to prematurely converge to suboptimal weight configurations. Additionally, it should encourage systematic exploration of the weight space to discover non-obvious adaptation strategies that static approaches cannot find.

Our reward function is defined as follows, which encourages both immediate performance improvements and strategic exploration of the weight space:
\begin{equation}
R_t = \alpha \cdot \frac{\Delta L_{\text{recon}}}{L_{\text{recon}}^{\text{prev}}} + \beta \cdot \Delta \text{Acc}_{\text{cls}} + \gamma \cdot B_{\text{significant}} + \delta \cdot B_{\text{entropy}}.
\end{equation}

The reward components serve distinct purposes:
\begin{itemize}
    \item Reconstruction Improvement ($\Delta L_{\text{recon}}/L_{\text{recon}}^{\text{prev}}$): Normalized loss reduction encourages consistent visual quality enhancement;
    \item Classification Gain ($\Delta \text{Acc}_{\text{cls}}$): Direct accuracy improvement rewards semantic preservation;
    \item Significance Bonus ($B_{\text{significant}}$): Additional reward for achieving improvements beyond a threshold (e.g., 5\% gain);
    \item Exploration Incentive ($B_{\text{entropy}}$): Entropy-based bonus encouraging diverse weight configurations during early training;
\end{itemize}

The significance bonus $B_{\text{significant}}$ is triggered when either task achieves substantial improvement beyond normal fluctuations:
\begin{equation}
B_{\text{significant}} = 
\begin{cases}
1.0, & \text{if } \frac{\Delta L_{\text{recon}}}{L_{\text{recon}}^{\text{prev}}} > 0.05 \text{ or } \Delta \text{Acc}_{\text{cls}} > 0.05 \\
0.0, & \text{otherwise}
\end{cases}.
\end{equation}

This fixed threshold provides consistent incentives for breakthrough improvements on either reconstruction or classification tasks, preventing the agent from settling for marginal gains.

The exploration entropy bonus is calculated based on the diversity of recent weight selections:
\begin{equation}
B_{\text{entropy}} = -\sum_{i} p_i \log p_i, \quad \text{where } p_i = \frac{\text{count}(\lambda_{\text{recon}}^{(i)})}{N_{\text{window}}}.
\end{equation}
Here, $\lambda_{\text{recon}}^{(i)}$ represents discretized weight bins from the last $N_{\text{window}} = 50$ actions, encouraging the agent to explore diverse weight configurations rather than repeatedly selecting similar values. The scaling factors $\{\alpha, \beta, \gamma, \delta\}$ are tuned to balance immediate performance gains with long-term strategic exploration.

\subsection{RL Network Architecture and Training}

\subsubsection{Policy Network Architecture}

We implement a Deep Q-Network (DQN) with careful architectural choices to ensure stable learning in our continuous action space. Key design decisions include:
\begin{itemize}
    \item LeakyReLU activation (negative slope = 0.01): Prevents dead neurons and improves gradient flow;
    \item Softplus output activation: Ensures strictly positive weights while maintaining smooth gradients;
    \item Hidden dimensions: [64, 32] neurons, balancing expressiveness with sample efficiency.
\end{itemize}

\subsubsection{Experience Replay and Target Networks}

To address the non-stationarity and correlation issues inherent in online RL, we employ:

\begin{itemize}
    \item Prioritized Experience Replay: We use a buffer of 10,000 transitions with TD (Temporal Difference)-error-based sampling probabilities, ensuring important experiences are revisited more frequently.
    \item Target Network: Separate Q-network updated via Polyak averaging:
    \begin{equation}
    \theta_{\text{target}} \leftarrow \tau \cdot \theta_{\text{policy}} + (1 - \tau) \cdot \theta_{\text{target}}, 
    \end{equation}
    with $\tau = 0.005$ for stable value estimation. This gradual Polyak averaging update ensures the target network parameters evolve slowly, smoothing out rapid fluctuations and preventing divergence during training.  
\end{itemize}

\subsubsection{Exploration Strategy}

Our exploration strategy combines $\epsilon$-greedy action selection with intelligent random sampling:
\begin{equation}
a_t = 
\begin{cases}
\pi_{\theta}(s_t), & \text{with probability } 1-\epsilon \\
a_{\text{random}}, & \text{with probability } \epsilon
\end{cases},
\end{equation} where $\epsilon$ decays from 1.0 to 0.05 over 50,000 steps. Here, $\pi_{\theta}(s_t)$ denotes the action chosen by the policy network, parameterized by $\theta$, when observing state $s_t$. Crucially, random actions $a_{\text{random}}$ are sampled from a mixture distribution:  

\begin{itemize}
    \item 70\% from uniform distribution over [0, 1]
    \item 30\% from a beta distribution favoring extreme values ($\alpha = \beta = 0.5$)
\end{itemize}
This encourages exploration of both balanced and specialized weight configurations.

\subsection{Integration with JSCC Framework}

The RL agent integrates with our JSCC training pipeline through the following mechanism:
\begin{itemize}
    \item State Observation: At each training step, the agent observes current channel conditions, recent performance metrics, and training progress;
    \item Weight Prediction: The policy network outputs task weights based on the observed state;
    \item Loss Computation: The multi-objective loss (Eq.~\eqref{eq:multiloss}) is computed using the RL-determined weights;
    \item Gradient Update: Both the JSCC model and RL agent are updated based on their respective objectives;
    \item Experience Storage: The transition $(s_t, a_t, r_t, s_{t+1})$ is stored for future learning.
\end{itemize}

This integration enables continuous adaptation throughout training and inference, with the system learning to anticipate and respond to changing conditions.

\subsection{Performance Adaptive Characteristics and Adaptation Rationale}

Our RL-based adaptive weighting exhibits several emergent behaviors that reflect fundamental optimization principles in semantic communication:

\begin{itemize}
    \item SNR-Aware Resource Allocation: The agent learns to emphasize reconstruction weights at low SNRs and shift toward classification weights at high SNRs, reflecting the principle that limited latent capacity must prioritize structural preservation when channel corruption is severe. At low SNRs, heavy noise destroys fine-grained semantic features regardless of emphasis, making basic visual structure the prerequisite for any subsequent understanding. Conversely, reliable high-SNR transmission enables aggressive allocation toward discriminative features that maximize classification performance;
    
    \item Content-Complexity Adaptation: For semantically complex images containing multiple objects or fine textures, the agent increases classification emphasis compared to simple scenes. This behavior emerges because complex scenes require more discriminative capacity to distinguish between competing semantic interpretations, while simple images with clear object-background separation can sacrifice semantic emphasis for structural fidelity without losing recognizability. The agent effectively learns that feature-rich content demands proportionally greater resource allocation for semantic preserving;
    
    \item Curriculum-Based Training Progression: Early training phases prioritize reconstruction, with weights gradually shifting toward a balanced allocation as encoder features improve. This trend reflects an implicit curriculum learning strategy: the untrained encoder initially yields poor latent representations, making reconstruction-focused learning essential for establishing meaningful feature extraction. As visual understanding improves, training shifts toward semantic tasks, enhancing discriminative capacity. This sequential specialization promotes faster convergence by progressively building semantic understanding on a foundation of robust representations;

    \item Optimization Efficiency: RL-guided training reaches target performance faster than grid-searched static weights by automatically discovering optimal weight trajectories that adapt to both training dynamics and instantaneous conditions, eliminating the need for expensive hyperparameter search across the multi-dimensional weight space.
\end{itemize}

These adaptation designs help our RL agent discover principled resource allocation strategies aligned with information-theoretic principles: preserving essential structure under severe constraints while maximizing task-specific performance when capacity permits. This automatic discovery of nuanced optimization strategies enables robust operation across the diverse conditions encountered in dynamic wireless environments.

\section{Module-Specific Low-Rank Adaptation}
\label{sec:CSLRA}

While LoRA enables parameter-efficient fine-tuning by introducing trainable low-rank matrices to model adaptations without modifying original weights, conventional approaches often adopt uniform LoRA configurations across all model components. This overlooks the diverse adaptation needs of heterogeneous modules. In our TOAST framework, architectural components such as Swin Transformer attention blocks, diffusion score networks, and classification heads demonstrate varying parameter sensitivities and distinct adaptation behaviors, necessitating module-specific tuning strategies.

\begin{figure}[t!]
  \centering
  \includegraphics[width=0.5\textwidth]{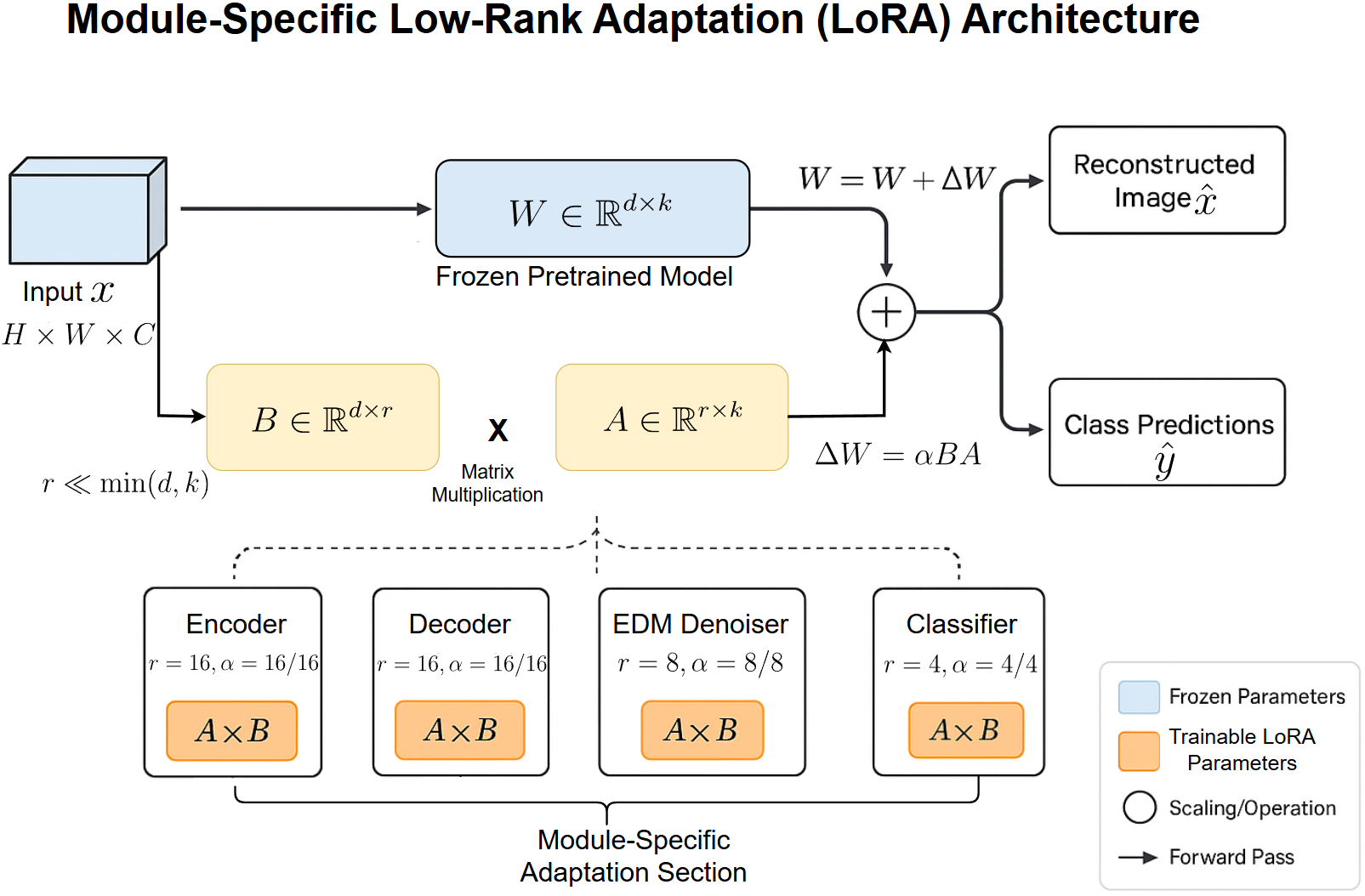}
  \caption{Module-Specific Low-Rank Adaptation (LoRA) architecture for the TOAST framework. The input tensor $x \in \mathbb{R}^{H \times W \times C}$ is processed by a frozen weight matrix $W \in \mathbb{R}^{d \times k}$ and a parallel LoRA path using $A \in \mathbb{R}^{r \times k}$ and $B \in \mathbb{R}^{d \times r}$, yielding $\Delta W = \alpha BA$ and adapted weight $W' = W + \Delta W$. The output supports image reconstruction $\hat{x}$ and classification $\hat{y}$. Each component (encoder, decoder, denoiser, classifier) uses task-specific LoRA ranks $r_c$ and scaling $\alpha_c = \hat{\alpha}_c / r_c$. Blue blocks denote frozen weights, orange blocks indicate trainable LoRA parameters, and circles represent operations.}

  \label{fig:lora}
\end{figure}

We propose a module-specific LoRA approach that tailors adaptation strategies to each architectural element. Rather than using identical rank and scaling parameters across all components, our approach recognizes that encoders require subtle feature extraction adjustments, decoders need substantial adaptation for corrupted latent reconstruction, diffusion models must recalibrate denoising schedules for channel-specific noise profiles, and classifiers benefit from minimal decision boundary adjustments. This heterogeneous adaptation strategy, illustrated in Fig.~\ref{fig:lora}, achieves superior channel adaptation performance with minimal parameter overhead compared to uniform LoRA approaches.

\subsection{Low-Rank Adaptation Foundations}

LoRA approximates full weight updates through low-rank matrix decomposition. For a pretrained weight matrix $W \in \mathbb{R}^{d \times k}$, the adapted weight becomes:

\begin{equation}
W' = W + \Delta W = W + BA,
\end{equation} where $B \in \mathbb{R}^{d \times r}$ and $A \in \mathbb{R}^{r \times k}$ with rank $r \ll \min(d, k)$. This decomposition reduces trainable parameters from $d \times k$ to $r \times (d + k)$, often achieving 100-1000× reduction.

Critically, LoRA does not merely "activate" dormant parameters. It introduces entirely new transformation pathways that specialize the frozen model for new conditions. The original model's capabilities remain intact, while the low-rank updates capture channel-specific adaptations.

\subsection{Enhanced Module-Specific LoRA Framework}

\subsubsection{Module-Specific Parameterization}

We introduce adaptation configurations for each component:
\begin{equation}
\Delta W_c = \alpha_c B_c A_c, \quad \text{with } \alpha_c = \frac{\hat{\alpha}_c}{r_c},
\end{equation} where $c \in \{\text{encoder}, \text{decoder}, \text{denoiser}, \text{classifier}\}$. The module-specific rank $r_c$ and scaled learning rate $\alpha_c$ enable precise capacity allocation.

\subsubsection{Initialization Strategy Differentiation}

Component functionality dictates initialization strategy:

\begin{itemize}
    \item Feature Extractors (encoder, denoiser): Kaiming uniform initialization \cite{he2015delv} is applied to the $A$ matrices to preserve variance through ReLU-like activations:
    \begin{equation}
    A_{ij} \sim \mathcal{U}\left(-\sqrt{\frac{6}{r_c}}, \sqrt{\frac{6}{r_c}}\right).
    \end{equation}
    
    \item Generative Components (decoder): initialization using a normal distribution with standard deviation $\sigma = 0.02$ encourages smooth generative adaptations;
    
    \item Discriminative Components (classifier): Xavier initialization \cite{glorot2010understanding} ensures balanced gradient flow during back propagation:
    \begin{equation}
    A_{ij} \sim \mathcal{U}\left(-\sqrt{\frac{6}{r_c + k}}, \sqrt{\frac{6}{r_c + k}}\right).
    \end{equation}
\end{itemize}
All $B$ matrices are initialized to zero, ensuring $\Delta W = 0$ initially and enabling a smooth, progressive departure from the pretrained model parameters.

\subsubsection{Specialized LoRA Integration Across Framework Modules}

\noindent\textbf{Window-Based Attention in Swin Transformer:} In our Swin Transformer blocks, we integrate LoRA adapters into the query–key–value projection matrices by setting their rank to $r_{\text{qkv}} = 2r_{\text{base}}$, which enables the model to capture more nuanced attention pattern shifts without bloating parameter count. The output projection layers, on the other hand, employ a more modest rank of $r_{\text{out}} = r_{\text{base}}$ to efficiently recombine features after attention operations. We leave the relative position bias frozen, ensuring that spatial encoding remains consistent and that LoRA updates focus solely on adapting the weight projections.

\noindent\textbf{Convolutional Layers:} To preserve spatial locality while adapting channel-wise interactions in convolutional components, we decompose each 2D convolutional weight update into a Kronecker-product form:
\begin{equation}
\Delta W_{\text{conv}} = B_{\text{spatial}} \otimes A_{\text{channel}},
\end{equation}
where $B_{\text{spatial}}$ captures spatial filtering adjustments and $A_{\text{channel}}$ encodes channel-level scaling. This formulation preserves the convolution’s structural inductive bias while delivering fine‐grained, low‐rank adaptations across both dimensions.

\textbf{Diffusion Score Networks in EDM:} The diffusion model in our EDM denoiser employs time-conditioned architectures that require specialized LoRA integration due to their dual dependency on spatial features and temporal noise schedules. Unlike standard neural networks, the diffusion score networks $s_\theta(\mathbf{z}_t, t)$ must simultaneously process corrupted latent features $\mathbf{z}_t$ and temporal conditioning information $t$ that encodes the current noise level in the denoising trajectory.

The core challenge lies in adapting both spatial processing pathways and temporal embedding mechanisms independently. Spatial features require adaptation to channel-specific corruption patterns, while temporal conditioning must adjust to different noise characteristics across varying channel types. We address this through module-specific LoRA placement:
\begin{align}
h &= \text{GroupNorm}(\text{Conv}_1(x) + \Delta W_1(x)), \\
y &= \text{Conv}_2(h + \text{MLP}(t) + \Delta W_t(t)) + \Delta W_2(h) + x,
\end{align} where $x \in \mathbb{R}^{C \times H \times W}$ represents the input corrupted latent, $t \in \mathbb{R}$ denotes the diffusion timestep encoding noise level, $h \in \mathbb{R}^{C \times H \times W}$ is the intermediate spatial representation, and $y \in \mathbb{R}^{C \times H \times W}$ represents the denoised output. The separate adapters serve distinct purposes: $\Delta W_1, \Delta W_2$ modify spatial feature processing to handle channel-specific corruption patterns, while $\Delta W_t$ adjusts temporal conditioning to match the noise characteristics of different channel types (e.g., bursty impulse noise vs. continuous fading). This decomposition enables independent optimization of spatial and temporal adaptation pathways while preserving the residual connection structure critical for diffusion model stability.

\subsection{Channel-Specific Adaptation Strategy}

Our strategy for adapting to previous unseen channel conditions comprises the following steps:

\begin{enumerate}
    \item Channel Profiling: Sample 1\% of the original data under new channel conditions to simulate a challenging low-data adaptation scenario;
    \item Adapter Initialization: Initialize module-specific LoRA modules with predetermined ranks tailored to each module's functional role;
    \item Rapid Fine-tuning: Efficiently adapt with up to 5 epochs, using early stopping to minimize overfitting and training time;
    \item Adapter Selection: Maintain a library of channel-specific adapters, allowing for instantaneous switching between configurations as channel conditions change.
\end{enumerate}

\subsection{Memory and Computational Efficiency}

Our module-specific LoRA approach achieves substantial efficiency gains while maintaining end-to-end model performance. Table~\ref{tab:lora_efficiency} quantifies these improvements in terms of resource-saving and adaptation time reduction.

\begin{table}[h]
\centering
\caption{LoRA adapter efficiency metrics for TOAST framework}
\label{tab:lora_efficiency}
\renewcommand{\arraystretch}{1.2}
\setlength\tabcolsep{0.6em}
\begin{tabular}{lccc}
\hline
\textbf{Metric} & \textbf{Full Model} & \textbf{LoRA} & \textbf{Reduction} \\
\hline
Parameters & 35.99M & 798.72K & 45.06× \\
Memory (FP16) & 72 MB & 1.6 MB & 45× \\
Adaptation Time & 2.5 hours & 3 minutes & 50× \\
\hline
\end{tabular}
\end{table}

These significant reductions facilitate rapid deployment in resource-constrained edge environments. The low inference overhead arises from the use of efficient low-rank matrix multiplication:
\begin{equation}
W'x = Wx + B(Ax),
\end{equation}
which requires only $r(d + k)$ additional operations compared to the $dk$ operations of full-rank multiplication, where $r \ll \min(d, k)$.

\subsection{Integration with System Components}

Module-specific LoRA modules integrate seamlessly throughout our architecture:

\begin{itemize}
    \item Swin Encoder/Decoder: Adapters are integrated into the attention projection layers and feed-forward networks within each Transformer block;
    \item EDM Denoiser: Adapters are applied to convolutional layers of ResNet blocks and time-embedding MLP projections;  
    \item Classification Head: A single adapter is employed in the final linear projection layer prior to the logit ouputs.
\end{itemize}

During inference, automatic channel detection triggers immediate adapter activation, enabling real-time specialization without manual intervention. This mechanism is essential for adapting to rapidly changing wireless environments.

Our module-specific LoRA strategy transforms channel adaptation from a computational bottleneck to a lightweight, fast process. By recognizing and exploiting the unique characteristics of each model component, we achieve superior adaptation quality with minimal resource requirements, which is a crucial capability for next-generation semantic communication systems operating in dynamic, heterogeneous wireless environments.

\section{Experimental Results and Comparative Analysis}
\label{sec:ERCA}

This section presents a comprehensive analysis of our TOAST framework, evaluating the effectiveness of three key components: the reinforcement learning-based task balancing mechanism, the EDM denoiser for latent space enhancement, and the module-specific LoRA adaptation for efficient channel-specific fine-tuning across diverse datasets and channel conditions.

\subsection{Datasets and Experimental Setup}

\subsubsection{Datasets}
We evaluate our system using standard image classification benchmarks that encompass various levels of complexity:

\begin{itemize}
    \item \textbf{SVHN}: The Street View House Numbers dataset contains over 600,000 32×32 color images of digits (0--9) collected from real-world scenes. It poses greater challenges than MNIST due to cluttered backgrounds and complex lighting. This dataset serves as the primary benchmark in our study, used across all experiments including adaptive weighting and LoRA adaptation.
    
    \item \textbf{CIFAR-10}: Comprising 60,000 32×32 color images across 10 classes (50,000 for training and 10,000 for testing), CIFAR-10 serves as a widely used benchmark for image classification and transmission tasks. We use it to benchmark our adaptive task balancing approach.

    \item \textbf{Intel Image Classification}: A real-world dataset containing natural images of six scene classes (e.g., forest, mountain, sea), resized to 32×32 resolution for consistency. It introduces diversity in content and texture, complementing SVHN and CIFAR-10.

    \item \textbf{MNIST}: A dataset of 70,000 grayscale handwritten digit images (28×28 pixels) from 10 classes. Despite its simplicity, MNIST is useful for observing performance trends in low-complexity scenarios and validating model generalization across modalities.
\end{itemize}

For all datasets, we apply standard augmentation techniques during training, including random horizontal flips and slight random rotations (where applicable). Images are normalized to the range [0, 1] before being fed into the model.

\subsubsection{Channel Models}

% As discussed in Section \ref{sec:PFRW}, we model wireless transmission using standard notation where $\mathbf{x}$ denotes the transmitted signal, $\mathbf{y}$ represents the received signal, and $\mathbf{n} \sim \mathcal{N}(0, \sigma^2)$ is additive white Gaussian noise with variance $\sigma^2$ determined by the signal-to-noise ratio (SNR). For fading channels, $\mathbf{h}$ represents the channel gain, while $\theta$ denotes phase distortion and $\mathbf{i}$ represents impulse interference.

For our main experiments on reinforcement learning agent and EDM ablation, we use the Additive White Gaussian Noise (AWGN) channel model: $y = x + n$, where the x is the transmitted signal, y is the received signal, and $n \sim \mathcal{N}(0, \sigma^2)$ denotes additive white Gaussian noise with variance, derived from the signal-to-noise ratio (SNR).

For our LoRA adaptation experiments, we evaluated four additional channel types to test adaptation capabilities:
    
\begin{itemize}
    \item \textbf{Rayleigh Fading}: Models wireless transmission with multipath effects, where the channel follows $y = hx + n$, where $h \sim \mathcal{CN}(0, 1)$, denotes the channel coefficients and $n \sim \mathcal{N}(0, \sigma^2)$ denotes additive white Gaussian noise.
    
    \item \textbf{Rician Fading}: Represents scenarios with line-of-sight components, modeled as $y = hx + n$, where $h$ follows a Rician distribution with K-factor of 2.
    
    \item \textbf{Phase Noise}: Introduces phase distortion with $y = xe^{j\theta} + n$, where $\theta \sim \mathcal{N}(0, \sigma_{\theta}^2)$ denotes phase distortion.
    
    \item \textbf{Impulse Noise}: Models bursty interference with $y = x + n + i$, where $i$ represents sparse high-magnitude impulses occurring with probability $p=0.01$.
\end{itemize}

In all experiments, we evaluate performance across an SNR range of 0-30 dB, with particular attention to the challenging low-SNR regime (0-10 dB).

\subsubsection{Implementation Details}

Our implementation is based on PyTorch and incorporates the following key components:

\begin{itemize}
    
    \item \textbf{Compression Rate:} In experiments largely deployed in our study (with SVHN and Cifar datasets), we compress $32 \times 32 \times 3 = 3{,}072$ input values into a latent representation of size $12 \times 12 \times 16 = 2{,}304$, resulting in a compression ratio of approximately 0.75. This corresponds to an effective bitrate of approximately 0.75 bits per pixel (bpp).

    \item \textbf{Training Configuration}: All models are trained using AdamW optimizer with a cosine learning rate schedule starting at 1e-4 and decaying to 1e-6 over 50 epochs. We use a batch size of 128 and apply gradient clipping at a maximum norm of 1.0.
    
    \item \textbf{Evaluation Metrics}: We assess system performance using two metric categories: reconstruction quality, measured by Peak Signal-to-Noise Ratio (PSNR) and Structural Similarity Index (SSIM), and semantic preservation quality, measured by classification accuracy and F1-score.
    
\end{itemize}

For the RL controller, we use a batch size of 64, discount factor $\gamma=0.99$, and soft update rate $\tau=0.005$. For LoRA adaptation, we employ module-specific ranks (encoder: 16, decoder: 16, EDM: 8, classifier: 4) with corresponding scaling factors.

\subsection{Architecture Comparison and Baseline Evaluation}

We begin our evaluation by comparing TOAST against alternative architectural approaches to establish the contribution of each design choice. Table~\ref{tab:architecture_comparison} presents a systematic comparison across different semantic communication architectures on the SVHN dataset under AWGN channel conditions.

\begin{table}[h]
\centering
\caption{Performance comparison of different architectures (PSNR in dB / Accuracy) on SVHN under AWGN channel conditions. Note: Swin Transformer corresponds to our JSCC-only baseline without EDM denoiser or RL adaptation.}
\label{tab:architecture_comparison}
\renewcommand{\arraystretch}{1.2}
\setlength\tabcolsep{0.1em}
\begin{tabular}{lccc}
\hline
\multirow{2}{*}{\textbf{Architecture}} & \multicolumn{3}{c}{\textbf{SNR}} \\
 & 5\,dB & 10\,dB & 15\,dB \\
\hline
CNN-based JSCC           & 12.8 / 48.7\% & 16.2 / 61.4\% & 20.9 / 71.6\% \\
CNN with DDPM            & 14.1 / 52.3\% & 17.8 / 65.2\% & 22.7 / 75.8\% \\
Swin Transformer & 15.3 / 55.2\% & 18.6 / 68.8\% & 24.3 / 78.1\% \\
\textbf{TOAST (Ours)}    & \textbf{23.7 / 65.0\%} & \textbf{27.7 / 80.3\%} & \textbf{32.1 / 88.9\%} \\
\hline
\end{tabular}
\end{table}

The results demonstrate a clear progression in performance capabilities. Traditional CNN-based JSCC \cite{bourtsoulatze2019deep} approaches, while providing a reasonable baseline, suffer from the inherent limitations of local receptive fields and limited long-range dependency modeling. The addition of DDPM-based denoising to CNN architectures (CNN with DDPM \cite{wu2024cddm}) provides moderate improvements, demonstrating that generative refinement can partially compensate for architectural limitations. However, the Swin Transformer architecture alone achieves superior performance over both CNN variants, highlighting the importance of attention mechanisms for capturing global dependencies in semantic communication tasks.

Moreover, our full TOAST framework achieves substantial performance enhancement over all baselines: 8.4 dB PSNR and 9.8\% accuracy improvement over the Swin Transformer alone, at 5 dB SNR, with consistent gains across all tested conditions. This shows the synergistic benefits of integrating hierarchical attention, diffusion-based denoiser, and reinforcement learning-driven task balancing within a unified framework.

\subsection{Training Convergence Analysis}

\begin{figure}[t!]
  \centering
  \includegraphics[width=0.5\textwidth]{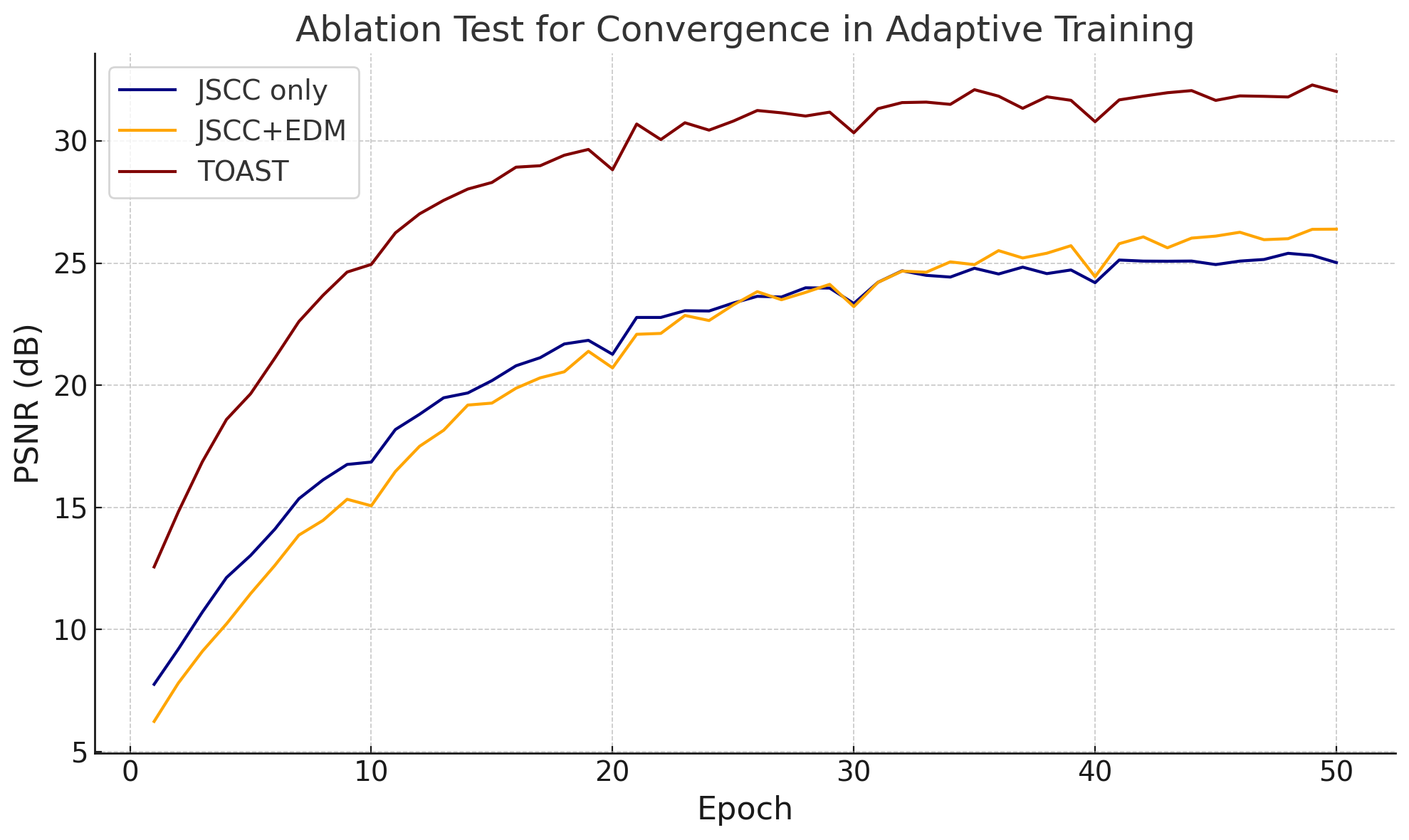}
  \caption{Training convergence analysis: PSNR evolution across epochs for different system configurations on SVHN dataset. TOAST demonstrates accelerated convergence and superior final performance compared to incremental baselines.}
  \label{fig:convergence_psnr}
\end{figure}

\begin{figure}[t!]
  \centering
  \includegraphics[width=0.5\textwidth]{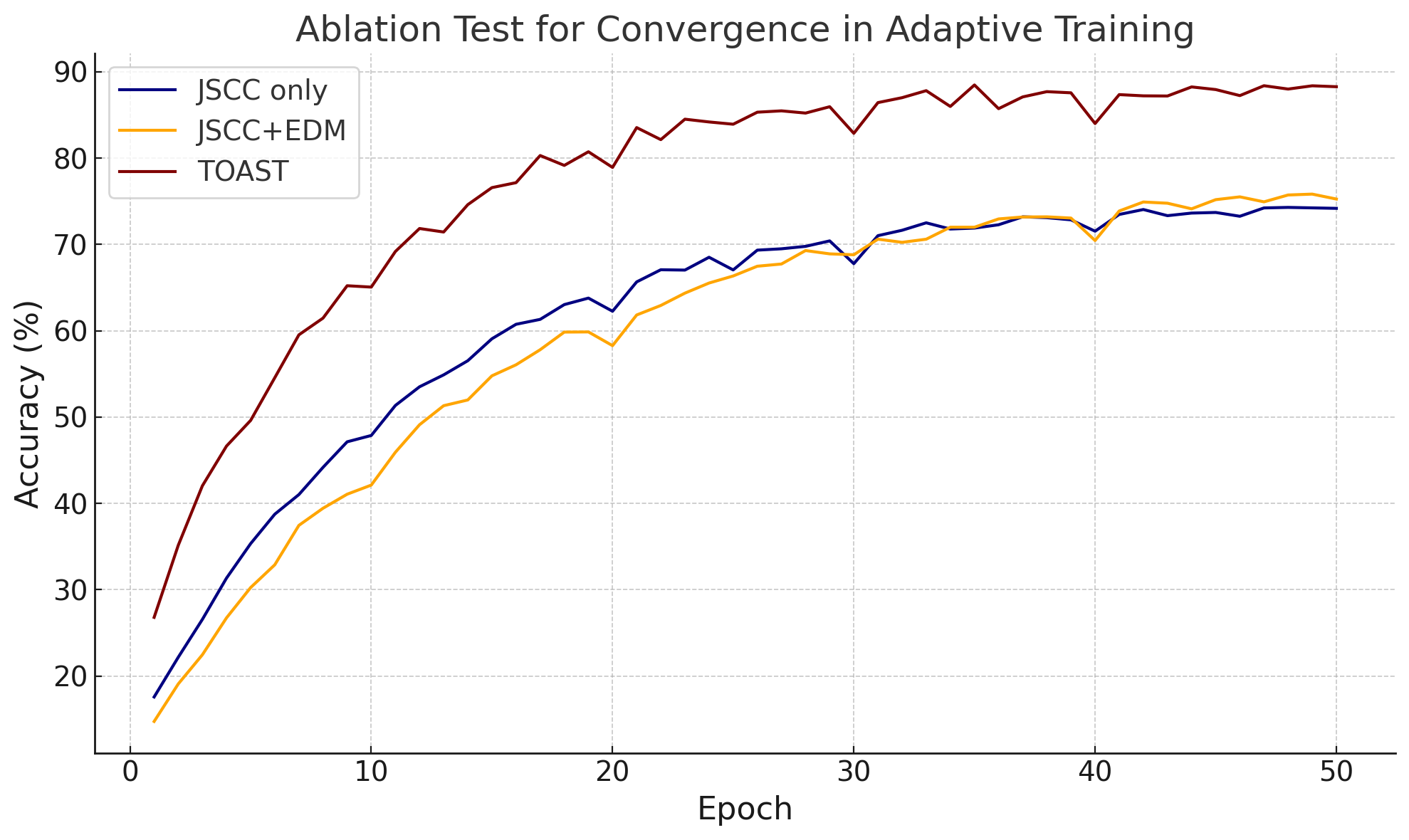}
  \caption{Training convergence analysis: Classification accuracy evolution across epochs for different system configurations on SVHN dataset. The RL-based adaptive weighting in TOAST enables faster convergence and better final accuracy.}
  \label{fig:convergence_accuracy}
\end{figure}

To gain deeper insight into the learning behavior of our proposed approach, we analyze the training convergence behavior across different system configurations. Fig.~\ref{fig:convergence_psnr} and Fig.~\ref{fig:convergence_accuracy} illustrate the evolution of reconstruction quality and classification accuracy throughout the training process.

The convergence analysis reveals several important insights. First, TOAST achieves rapid early convergence, reaching competitive performance within the first 10 epochs, significantly faster than other approaches. The JSCC-only baseline shows steady but slow improvement, reaching approximately 25 dB PSNR and 75\% accuracy after 50 epochs. Incorporating the EDM denoiser (JSCC+EDM) yields moderate improvements while maintaining similar convergence characteristics.

In contrast, TOAST demonstrates markedly different learning dynamics. The RL-based adaptive task balancing enables the system to quickly discover effective weight configurations, leading to rapid performance gains in the initial training phases. By epoch 15, TOAST already surpasses the final performance of both baseline approaches, ultimately converging to approximately 32 dB PSNR and 88\% classification accuracy. This accelerated convergence is particularly valuable for practical deployment scenarios where training time and computational resources are constrained.

\subsection{Comprehensive Multi-Dataset Evaluation}

Table~\ref{tab:jscc_variants} presents a comprehensive evaluation across four diverse image datasets, demonstrating the generalization capabilities of our approach across varying data characteristics and semantic complexities.

\begin{table}[h]
\centering
\caption{Performance of JSCC variants (PSNR in dB / Accuracy) under AWGN channel conditions.}
\label{tab:jscc_variants}
\renewcommand{\arraystretch}{1.0}
\setlength\tabcolsep{0.5em}
\begin{tabular}{lccc}
\hline
\multirow{2}{*}{\textbf{Model}} & \multicolumn{3}{c}{\textbf{SNR}} \\
 & 5\,dB & 10\,dB & 15\,dB \\
\hline
Dataset & \multicolumn{3}{c}{\textbf{SVHN}} \\
JSCC only            & 15.3 / 55.2\% & 18.6 / 68.8\% & 24.3 / 78.1\% \\
JSCC+EDM             & 20.7 / 58.3\% & 24.1 / 75.6\% & 28.5 / 84.2\% \\
\textbf{TOAST}       & \textbf{23.7 / 65.0\%} & \textbf{27.7 / 80.3\%} & \textbf{32.1 / 88.9\%} \\
\hline
Dataset & \multicolumn{3}{c}{\textbf{CIFAR-10}} \\
JSCC only            & 14.8 / 38.4\% & 17.9 / 55.2\% & 22.6 / 65.8\% \\
JSCC+EDM             & 20.2 / 42.1\% & 23.4 / 60.8\% & 26.8 / 71.5\% \\
\textbf{TOAST}       & \textbf{23.2 / 48.8\%} & \textbf{27.0 / 66.5\%} & \textbf{30.4 / 77.2\%} \\
\hline
Dataset & \multicolumn{3}{c}{\textbf{INTEL IMAGE}} \\
JSCC only            & 14.1 / 35.2\% & 15.1 / 65.8\% & 19.8 / 72.1\% \\
JSCC+EDM             & 15.7 / 48.9\% & 17.6 / 67.6\% & 21.0 / 75.6\% \\
\textbf{TOAST}       & \textbf{19.8 / 63.4\%} & \textbf{22.2 / 75.3\%} & \textbf{25.6 / 82.3\%} \\
\hline
Dataset & \multicolumn{3}{c}{\textbf{MNIST}} \\
JSCC only            & 26.7 / 78.4\% & 30.5 / 91.2\% & 34.1 / 95.8\% \\
JSCC+EDM             & 31.2 / 82.1\% & 36.0 / 93.9\% & 38.3 / 97.4\% \\
\textbf{TOAST}       & \textbf{34.2 / 88.8\%} & \textbf{39.3 / 97.6\%} & \textbf{41.9 / 98.9\%} \\
\hline
\end{tabular}
\end{table}

This cross-dataset evaluation reveals several important patterns. Performance scales predictably with dataset complexity: MNIST (comprising simple grayscale digits) achieves the highest absolute performance, followed by Intel Image (with 6-class natural scenes), SVHN (comprising real-world digits with complex backgrounds), and CIFAR-10 (containing 10-class natural objects with the greatest visual diversity). Importantly, TOAST maintains consistent relative improvements across all datasets, demonstrating robust generalization capabilities.

The improvement patterns are particularly noteworthy. Adding EDM denoising typically provides 5-6 dB gain in PSNR and 3-5\% gain in classification accuracy across datasets, while the complete TOAST framework delivers an additional gain in 3-4 dB PSNR and an extra 5-7\% gain in classification accuracy. These consistent improvements across diverse data characteristics confirm the effectiveness of our approach across a wide range of visual and semantic contexts rather than dataset-specific artifacts.

\subsection{EDM Denoiser Ablation Study}

To isolate the contribution of our EDM denoiser component, we conduct comprehensive ablation studies examining both qualitative and quantitative improvements.

\subsubsection{Visual Quality Enhancement}

\begin{figure}[t!]
  \centering
  \includegraphics[width=0.5\textwidth]{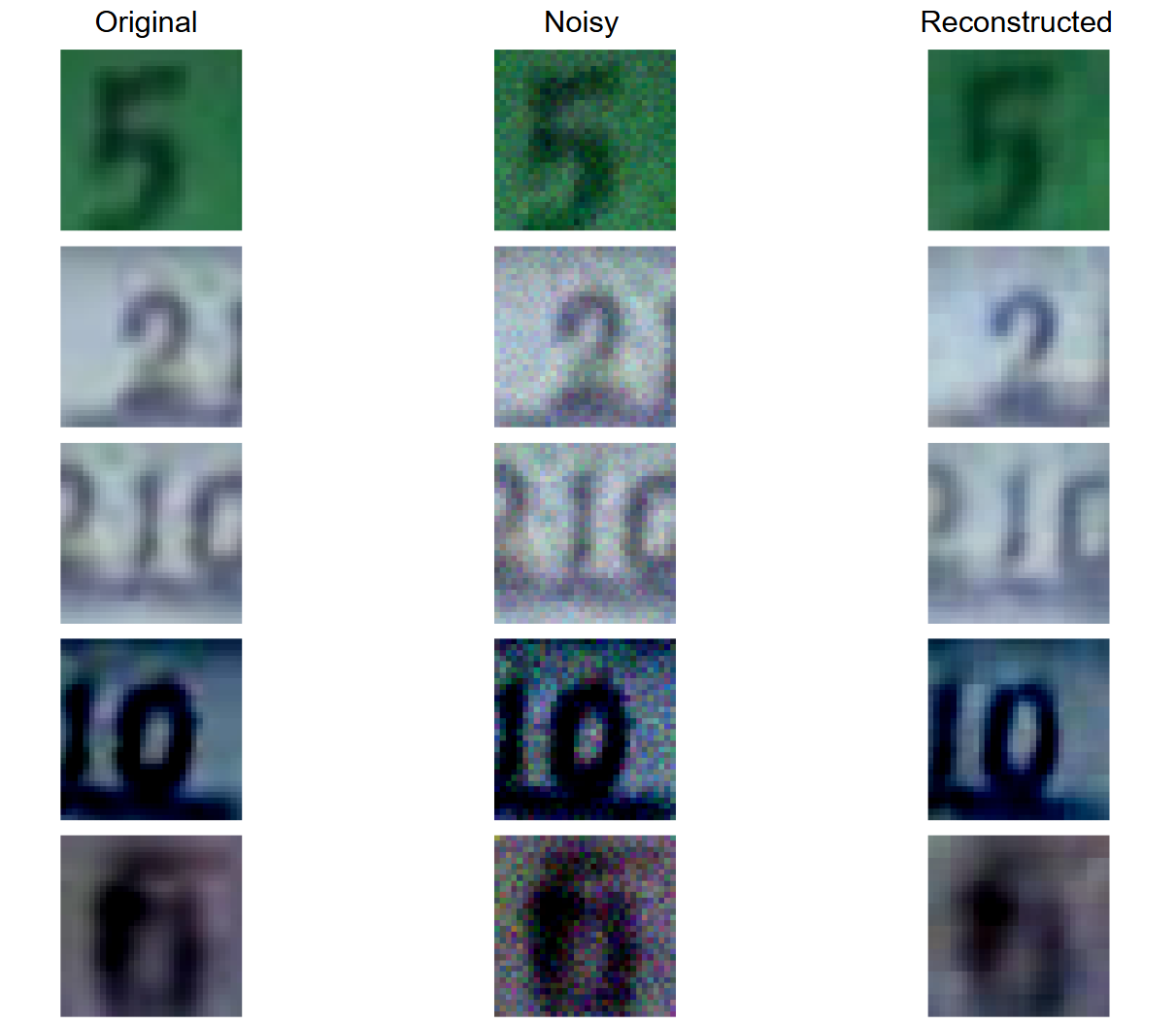}
  \caption{Visual comparison of reconstruction quality on SVHN dataset at 15 dB SNR under AWGN channel. Left column shows original images, middle column shows corrupted versions after channel transmission, and right column displays reconstructed outputs using our TOAST framework. The EDM denoiser effectively recovers structural details and semantic content essential for digit recognition.}
  \label{fig:visual_comparison}
\end{figure}

Fig.~\ref{fig:visual_comparison} demonstrates the visual quality improvements achieved by our EDM denoiser through representative examples from the SVHN dataset under challenging noise conditions.

The visual comparison reveals the EDM denoiser's effectiveness in recovering critical structural and semantic information from severely corrupted latent representations. Original digit images maintain clear structure and recognizable semantic content, while channel-corrupted versions suffer significant degradation with visible noise artifacts that would impede both human recognition and automated classification. TOAST reconstruction successfully removes channel artifacts while preserving essential digit characteristics, demonstrating the denoiser's ability to distinguish between meaningful signal variations and noise-induced corruptions.

\subsubsection{Quantitative Performance Analysis}

\begin{figure}[t!]
  \centering
  \includegraphics[width=0.5\textwidth]{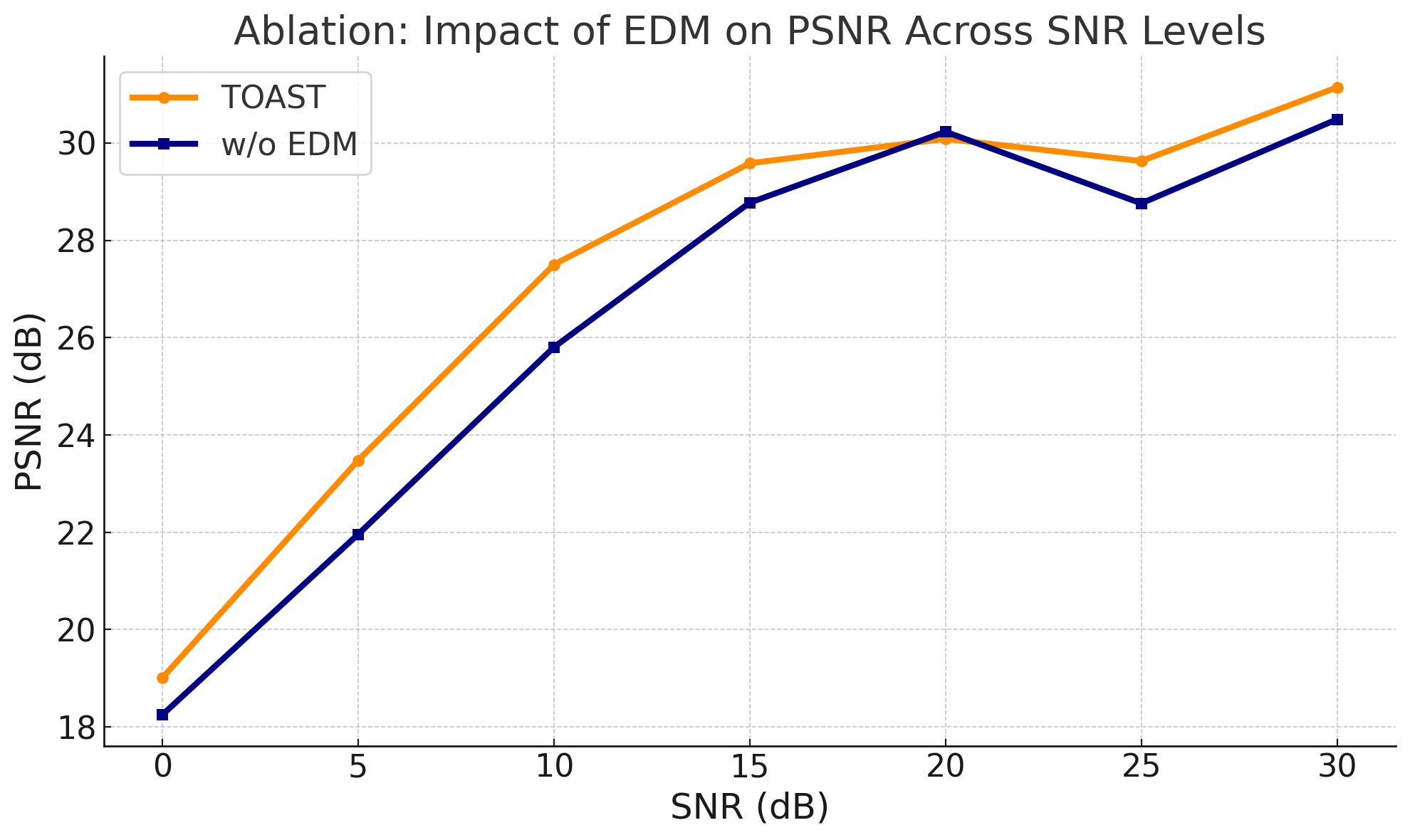}
  \caption{PSNR performance comparison between TOAST and TOAST without EDM across SNR levels on SVHN dataset under AWGN channel. The EDM denoiser provides substantial improvements particularly at low SNR conditions, with the largest gains in challenging noise environments.}
  \label{fig:edm_psnr_ablation}
\end{figure}

\begin{figure}[t!]
  \centering
  \includegraphics[width=0.5\textwidth]{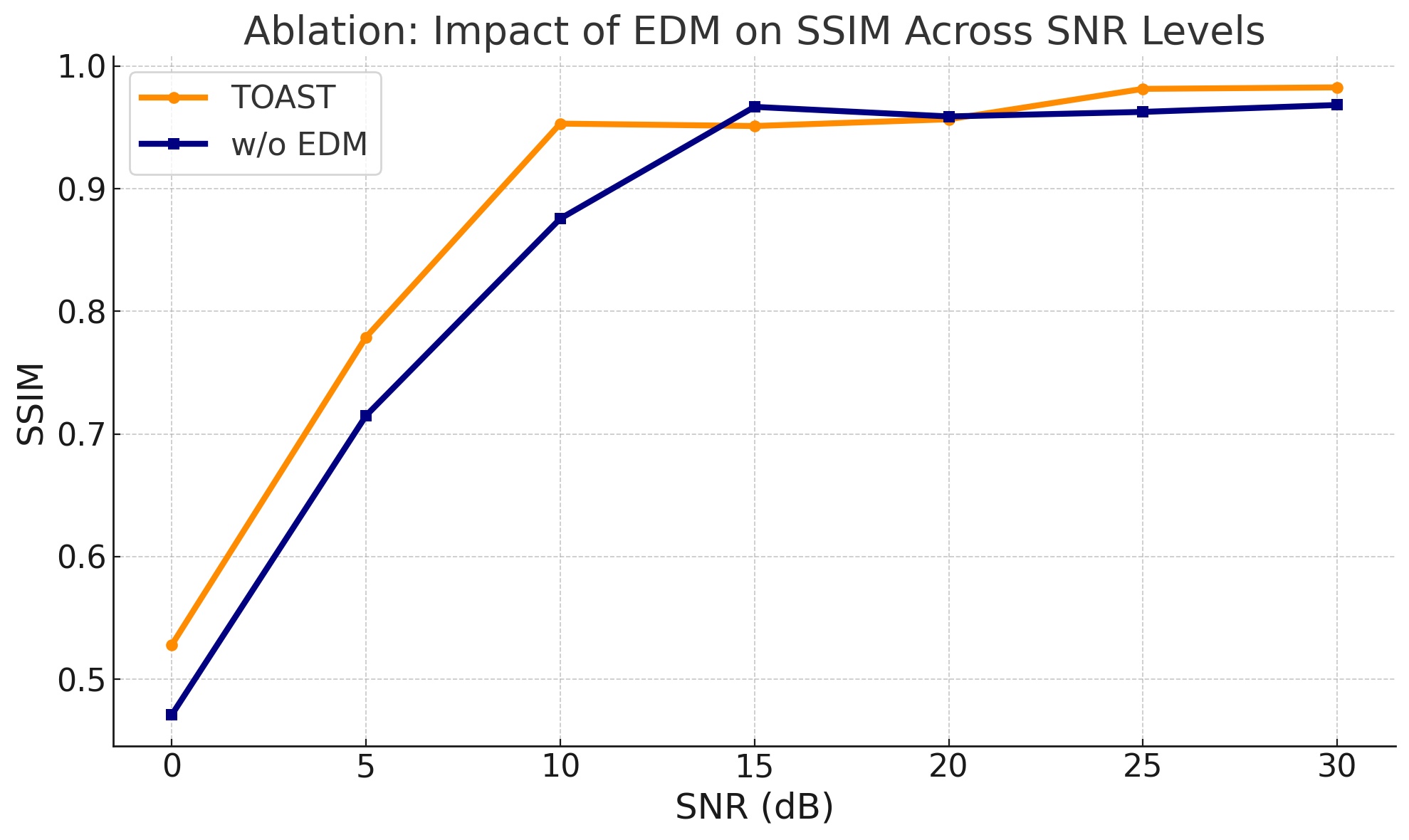}
  \caption{SSIM performance comparison between TOAST and TOAST without EDM across SNR levels on SVHN dataset under AWGN channel. The structural similarity improvements are most pronounced at low SNR levels, indicating the EDM's effectiveness in preserving perceptually important image structure.}
  \label{fig:edm_ssim_ablation}
\end{figure}

Fig.~\ref{fig:edm_psnr_ablation} and Fig.~\ref{fig:edm_ssim_ablation} present a comprehensive quantitative assessment of the EDM denoiser's impact across the full operational SNR range.

The quantitative ablation reveals consistent and substantial benefits from EDM integration. For PSNR performance, the denoiser provides the most significant improvements under low SNR conditions, where channel noise presents the greatest challenge. At 0 dB SNR, EDM contributes approximately 1.2 dB of improvement, while at 5 dB SNR, the gain remains substantial at nearly 1.5 dB. Although the improvements gradually decrease at higher SNRs, they remain appreciable even at 30 dB, demonstrating the denoiser's effectiveness across the entire operational range.  

Although the PSNR curve shows a small 0.3 dB dip at 25 dB, this reflects finite-sample noise variability coupled with the RL agent’s brief reprioritization toward semantic accuracy. In that narrow SNR window, the policy increases the classification weight—improving downstream task performance while marginally trading off raw reconstruction fidelity. This intentional, transient behavior underscores our task-driven design and does not weaken overall robustness. The perceptual gains measured by SSIM remain fully consistent at this point, further illustrating the EDM denoiser’s stable benefit.

The SSIM analysis provides complementary insights into perceptual quality preservation. At low SNRs (0-5 dB), EDM delivers dramatic structural similarity improvements, with SSIM gains of 0.06-0.08 units. These improvements reflect the denoiser's ability to recover global image structure and maintain spatial relationships critical for human perception and downstream semantic tasks. At higher SNRs, SSIM improvements become more modest but remain consistent, demonstrating EDM's capability to enhance fine detail preservation without introducing artifacts.

The convergence of both PSNR and SSIM metrics at high SNRs confirms that EDM operates as intended: providing substantial enhancement under challenging conditions while gracefully reducing its intervention as channel quality improves. This adaptive behavior aligns with our design goals of robust operation across dynamic wireless environments.

\subsection{Classification Performance Enhancement}

Beyond reconstruction fidelity, our evaluation shows significant improvements in semantic task performance. Across all tested datasets and SNR conditions, TOAST consistently outperforms baseline approaches in classification accuracy. The improvements are most pronounced at low SNR conditions where channel noise severely degrades semantic information: typically 10-15\% accuracy gains at 5 dB SNR, reducing to 3-5\% gains at higher SNRs. This pattern confirms that our integrated approach successfully preserves task-relevant information under challenging transmission conditions while optimizing resource allocation as channel quality improves.

The experimental results collectively demonstrate that TOAST achieves superior performance across multiple evaluation dimensions: architectural innovation, training efficiency, cross-dataset generalization, and module-specific contributions. These collective strengths position TOAST as an effective solution for practical semantic communication systems operating in dynamic wireless environments.

\subsection{LoRA Fine-tuning Experiments and Results}

This subsection shows that, beyond AWGN, LoRA supports rapid, parameter-efficient adaptation to varied channel impairments using minimal training data.

\subsubsection{Adaptation to Unseen Channel Conditions}

Table~\ref{tab:lora_adaptation} presents the performance improvement achieved through LoRA fine-tuning when adapting a model trained on AWGN channels to different channel impairments. The base model was pre-trained on AWGN channels with SNR ranging from 0--30 dB, then fine-tuned using module-specific LoRA adapters for each target channel condition.

\begin{table}[h]
\centering
\caption{Performance improvement from LoRA fine-tuning when adapting to different channel conditions on SVHN at 10 dB SNR. (1\% data, 5 epochs)}
\label{tab:lora_adaptation}
\renewcommand{\arraystretch}{1.2}
\setlength\tabcolsep{0.6em}
\begin{tabular}{lcccc}
\hline
\multirow{2}{*}{\textbf{Channel Type}} & \multicolumn{2}{c}{\textbf{Base Model}} & \multicolumn{2}{c}{\textbf{LoRA Fine-tuned}} \\
 & PSNR & Accuracy & PSNR & Accuracy \\
\hline
AWGN          & 27.73 & 87.66\% & 29.92 & 89.55\% \\
Rayleigh      & 11.36 & 28.92\% & 19.25 & 68.45\% \\
Rician        & 13.78 & 39.45\% & 24.76 & 74.82\% \\
Phase Noise   & 12.45 & 29.89\% & 21.67 & 72.33\% \\
Impulse Noise & 11.26 & 22.89\% & 18.67 & 62.74\% \\
\hline
\end{tabular}
\end{table}

\subsubsection{Parameter Efficiency Analysis}

Table~\ref{tab:lora_params} summarizes the LoRA adapter parameter breakdown and efficiency. The results highlight the significant parameter reduction achieved through our module-specific LoRA approach.

\begin{table}[h]
\centering
\caption{LoRA adapter parameter breakdown.}
\label{tab:lora_params}
\renewcommand{\arraystretch}{1.2}
\setlength\tabcolsep{1.3em}
\begin{tabular}{lrrr}
\hline
\multirow{2}{*}{\textbf{Component}} & \textbf{Original} & \textbf{LoRA} & \textbf{\% of} \\
                                    & \textbf{Params}   & \textbf{Params} & \textbf{Orig.} \\
\hline
Encoder    & 12.54M & 460.80K & 1.47\%  \\
Decoder    & 7.14M  & 337.92K & 2.58\%  \\
EDM        & 16.21M & 122.88K & 0.76\%  \\
Classifier & 0.10M  & 6.22K   & 6.15\%  \\
\hline
\textbf{Total model} & 35.99M & 798.72K & 2.22\% \\
\hline
\end{tabular}
\end{table}

Our module-specific LoRA adaptation approach achieves significant performance improvements with minimal parameter overhead, reducing the total trainable parameters by a factor of 45.06×. This demonstrates the effectiveness of our approach for efficient adaptation to diverse channel conditions, enabling rapid deployment in dynamic wireless environments.

In summary, the experimental results demonstrate that the TOAST framework, which integrates a Swin Transformer JSCC backbone, an EDM denoiser, reinforcement learning-based task balancing, and module-specific LoRA adaptation, achieves robust performance across diverse channel conditions and datasets. The reinforcement learning controller dynamically prioritizes tasks based on channel quality, while the LoRA adapters facilitate efficient fine-tuning for specific channel impairments. This leads to significant improvements in both reconstruction quality and semantic accuracy, highlighting the system’s effectiveness in dynamic wireless environments.

\section{Discussion, Limitations, and Conclusion}
\label{sec:DLC}

This paper introduces a novel TOSC architecture that integrates a Swin Transformer-based JSCC backbone, diffusion-based generative refinement, a reinforcement learning-driven multi-task scheduler, and module-specific LoRA adaptation. By combining these techniques, the proposed TOSC system achieves robust dual-task performance, delivering both high-fidelity reconstruction and accurate classification under varied and dynamic wireless conditions. This design provides a foundation for the development of adaptive, efficient, and semantically aware transmission strategies tailored to the needs of next-generation communication networks.

Experimental results confirm that dynamic task balancing via reinforcement learning consistently outperforms static weighting schemes across a wide range of SNR conditions. The system automatically emphasizes reconstruction when operating under low SNR and shifts toward semantic preservation as SNR improves. The diffusion-augmented decoder contributes to enhanced perceptual quality, particularly under severe noise conditions. Meanwhile, LoRA modules enable efficient adaptation to a broad range of channel impairments, including Rayleigh fading, Rician fading, phase distortion, and impulse interference, while requiring only a small proportion of trainable parameters. These components together form a cohesive and adaptive SemCom system that remains resilient as network conditions evolve. TOAST is ideally suited for heterogeneous 6G deployments— from autonomous‐vehicle networks that demand real‐time scene understanding to bandwidth‐constrained IoT systems—thanks to its adaptability in mobile edge environments where channel conditions and network topologies change frequently.

However, several limitations remain: the full TOAST model (Swin Transformer plus diffusion decoder) is still too heavy for highly resource-constrained devices; evaluation on standard vision benchmarks (SVHN, CIFAR, Intel Image, MNIST variants) may not reflect real-world diversity; simulated channels (AWGN, fading, impulse noise) omit complex effects like mobility-induced Doppler shifts and large-scale interference; and the RL-based adaptive weighting demands careful reward scaling and exploration tuning to ensure stable convergence in unforeseen conditions.

\bibliographystyle{IEEEtran}
\bibliography{IEEEabrv,bibi}

\end{document}